\documentclass{ecai} 
\usepackage{latexsym}
\usepackage{amssymb}
\usepackage{amsmath}
\usepackage{amsthm}
\usepackage{booktabs}
\usepackage{enumitem}
\usepackage{graphicx}
\usepackage{color}

\newcommand{\BibTeX}{B\kern-.05em{\sc i\kern-.025em b}\kern-.08em\TeX}
\usepackage{latexsym}
\usepackage{amsmath}
\usepackage{amsthm}
\usepackage{booktabs}
\usepackage{enumitem}
\usepackage{graphicx}
\usepackage{color}
\usepackage{subfig}

\usepackage{amsfonts}
\usepackage{xcolor}
\usepackage{xspace}

\newcount\Comments  % 0 suppresses notes to selves in text
\long\def\roie#1{{\ifnum\Comments=1\color{red} [Roie: #1]\fi}}
\long\def\arseni#1{{\ifnum\Comments=1\color{red} #1\fi}}
\long\def\ariel#1{{\ifnum\Comments=1\color{red} [Ariel: #1]\fi}}
\long\def\roni#1{{\ifnum\Comments=1\color{red} [Roni: #1]\fi}}

\newcommand{\ta}{TA$^*$\xspace}
\newcommand{\tapf}{TA$^*$+APF}
\newcommand{\sippsapf}{SIPPS+APF\xspace}
\newcommand{\pibtapf}{PIBT+APF\xspace}
\newcommand{\maxdist}{\ensuremath{d_{max}}}
\newcommand{\shape}{\ensuremath{\gamma}}
\newcommand{\astar}{A$^*$}

\begin{document}

%%%%%%%%%%%%%%%%%%%%%%%%%%%%%%%%%%%%%%%%%%%%%%%%%%%%%%%%%%%%%%%%%%%%%%%%

\begin{frontmatter}

%%% Use this command to specify your submission number.
%%% In doubleblind mode, it will be printed on the first page.

\paperid{7341} 

%%% Use this command to specify the title of your paper.

\title{Enhancing Lifelong Multi-Agent Path-finding by Using \\ Artificial Potential Fields}

%%% Use this combinations of commands to specify all authors of your 
%%% paper. Use \fnms{} and \snm{} to indicate everyone's first names 
%%% and surname. This will help the publisher with indexing the 
%%% proceedings. Please use a reasonable approximation in case your 
%%% name does not neatly split into "first names" and "surname".
%%% Specifying your ORCID digital identifier is optional. 
%%% Use the \thanks{} command to indicate one or more corresponding 
%%% authors and their email address(es). If so desired, you can specify
%%% author contributions using the \footnote{} command.

\author[]{\fnms{Arseniy}~\snm{Pertzovsky}}
\author[]{\fnms{Roni}~\snm{Stern}}
\author[]{\fnms{Ariel}~\snm{Felner}} 
\author[]{\fnms{Roie}~\snm{Zivan}} 

\address[]{Ben-Gurion University of the Negev}
% \address[B]{Short Affiliation of Second Author and Third Author}
% \address[C]{Short Alternate Affiliation of Third Author}

%%% Use this environment to include an abstract of your paper.

\begin{abstract}
We explore the use of Artificial Potential Fields (APFs) to solve Multi-Agent Path Finding (MAPF) and Lifelong MAPF (LMAPF) problems. In MAPF, a team of agents must move to their goal locations without collisions, whereas in LMAPF, new goals are generated upon arrival. We propose methods for incorporating APFs in a range of MAPF algorithms, including Prioritized Planning, MAPF-LNS2, and Priority Inheritance with Backtracking (PIBT). Experimental results show that using APF is not beneficial for MAPF but yields up to a 7-fold increase in overall system throughput for LMAPF. 
\end{abstract}

\end{frontmatter}

%%%%%%%%%%%%%%%%%%%%%%%%%%%%%%%%%%%%%%%%%%%%%%%%%%%%%%%%%%%%%%%%%%%%%%%%
%%%%%%%%%%%%%%%%%%%%%%%%%%%%%%%%%%%%%%%%%%%%%%%%%%%%%%%%%%%%%%%%%%%%%%%%
%%%%%%%%%%%%%%%%%%%%%%%%%%%%%%%%%%%%%%%%%%%%%%%%%%%%%%%%%%%%%%%%%%%%%%%%
\section{Introduction}

% APFs
\emph{Artificial Potential Fields (APFs)}~\cite{khatib1986potential} can be used to improve the deployment of mobile agents in an environment with obstacles. 
When using this physics-inspired approach, each agent is considered a particle that is moved on the basis of the attraction and repulsion forces that are applied to it. %Typically, % \ariel{in single agent search}, \roni{no, also in multi agent when using APFs}
Typically, obstacles and other agents generate repulsive forces, while targets and destination goals apply attractive forces~\cite{koren1991potential}. 
APFs have been used to solve many motion planning problems~\cite{mac2016heuristic,hagelback2009multi,daily2008harmonic}, with successful applications in obstacle avoidance of an unmanned aircraft~\cite{rezaee2012adaptive}, collision avoidance systems for automated vehicles~\cite{wahid2017study} and more~\cite{mac2016heuristic}.

Multi-agent Pathfinding (MAPF) is the problem of finding paths for a group of agents such that they reach their goals without colliding with each other~\cite{stern2019mapf}. 
Instances of MAPF exist in robotics~\cite{bartak2019multi}, automated warehouses~\cite{wurman2008coordinating,salzman2020research}, digital entertainment~\cite{ma2017feasibility} and more~\cite{morris2016planning}. 
%%%A range of algorithms for solving MAPF problems have been proposed over the years~\cite{surynek2016efficient,felner2017search,ShardSystems_Li_Koenig_2022,li2022mapf}, following different approaches and providing different solution quality guarantees. 
% MAPF is hard when centralized and coupled. We decouple and use Temporal A*. This causes congensions
Solving MAPF is computationally difficult since the number of possible actions at every time point is exponential in the number of agents.
In fact, for common optimization criteria, finding an optimal solution for MAPF is NP-hard~\cite{yu2013structure,surynek2010anOptimization}. 
Popular optimal and suboptimal algorithms, such as Prioritized Planning (PrP)~\cite{prp_2015}, 
Conflict-Based Search (CBS)~\cite{CBS_2015}, 
and Large Neighborhood Search for MAPF (MAPF-LNS)~\cite{li2021anytime}, 
mitigate this combinatorial challenge 
%to some extent the combinatorial challenge of planning for multiple agents 
by planning a path for each agent individually and imposing different forms of constraints to avoid collisions. 
%%\ariel{Do we want to say that they all have high levels and low levels and we work on the low levels which plan for a single agent only. AF2: I agree}\roni{seems out of context to me  here, but whatever works}
% Alternatively, algorithms such as PIBT~\cite{okumura2022priority}, LaCAM~\cite{okumura2023lacam}, and LaCAM$^*$~\cite{okumura2023lacam}, directly 
Alternatively, algorithms such as PIBT, LaCAM, and LaCAM$^*$~\cite{okumura2022priority,okumura2023lacam,okumura2023lacam}, directly search the space of \emph{configurations}, where a configuration is a set of all agents' locations in a given time step.

However, using existing algorithms often results in congestion occurring in areas of the environment that are shared by the shortest paths of multiple agents.
This congestion problem is exacerbated when solving {\em lifelong MAPF} (LMAPF)~\cite{li2021anytime,vsvancara2019online}, where an agent receives a new goal location whenever it reaches its current goal location. 
In LMAPF, congested areas tend to become more and more congested over time, resulting in a decrease in overall system efficiency.
The use of APFs is a natural approach to encourage agents to avoid congested areas, adding repulsion forces not only to avoid obstacles but also to avoid the paths of other agents. In this paper, we explore the potential of this approach.

% \begin{figure}[t]
% %\vspace{-8pt}
%    % \hspace{2pt}
%  \begin{center}
%   \includegraphics[scale=0.19]{pics/noPF_withPF2.png}
% % \vspace{-10pt}
%  \caption{(a) initial setting, arrows point to the goal locations; (b) a run without the usage of Artificial Potential Fields - congestion arises; (c) a run with the usage of Artificial Potential Fields (bigger circles) - agents choose alternative paths}
%  \label{Fig:with_without_pfs}
%  % \vspace{-18pt}
% \end{center}
% \end{figure}
% APF fails, need planning

%This is our main focus of the paper.
% This is demonstrated in Figure \ref{Fig:with_without_pfs}. Figure \ref{Fig:with_without_pfs}(a) shows the initial positions of 7 agents and the arrows point to the directions of their goal locations. Figure \ref{Fig:with_without_pfs}(b) demonstrates the congestion that may arise with a classical MAPF solver where all agents try to take the shortest path, therefore trying to avoid many constraints created by other agents. 
% Figure \ref{Fig:with_without_pfs}(c) represents a case where agents use APFs to maintain distance from each other by picking alternative paths, therefore confronting fewer constraints.
%\arseni{I've removed the picture that describes the idea of APFs. No one liked it.}
We first investigate using APFs to solve MAPF problems by following a simple, myopic approach: each agent plans its next move based on the forces of APFs of the other agents.
%where an agent's goal has an APF that attracts it and all other agents have an APF that repulses it.
The resulting MAPF algorithm (Direct APF) compares poorly with existing MAPF algorithms on standard MAPF benchmarks.
% \ariel{What are non-trivial problems} \arseni{I don't understand this sentence.} \roie{I revised it}
Apparently, relying only on attraction and repulsion forces in MAPF is too myopic, and longer-horizon path planning is needed. 
We, therefore, explore the addition of APFs into existing MAPF solvers. 
We integrate APFs into \emph{Temporal \astar\ }(\ta)~\cite{prp_2015} and \emph{Safe Interval Path
Planning with Soft constraints} (SIPPS)~\cite{li2022mapf}, which are single-agent pathfinding algorithms that plan for individual agents under different types of constraints.  
We modify the cost of agents' actions to consider the APFs created by other agents who have already planned their paths.
Consequently, the path returned for an agent not only optimizes fast arrival at the goal but also avoids areas congested by other agents. We also integrate APFs in a similar way in PIBT, LaCAM, and LaCAM$^*$. 
% We then consider PIBT~\cite{okumura2022priority}, LaCAM~\cite{okumura2023lacam}, and LaCAM$^*$~\cite{okumura2023lacam}, which are MAPF algorithms that search the space of \emph{configurations}, where 
% a configuration is a set of all agents' locations in a given time-step. 
Experimental results on a range of standard MAPF benchmark problems show that
our APF-augmented algorithms are not effective for classical MAPF. 
However, they are very effective when solving LMAPF, yielding up to a 7-fold increase in throughput. 

\section{Definition and Background}

% - MAPF
% - LMAPF
% - opt
% - sub opt
% - high level low level
% - TA*
% - PrP
% - LNS2
% - iStay

% \subsection{MAPF} no need
A classic MAPF problem is defined by a tuple $\langle k, G, s, g \rangle$ where
$k$ is the number of agents, 
$G = (V, E)$ is an undirected graph, 
$s: [1, ... , k] \rightarrow V$ maps an agent to a start vertex,
and $g: [1, ... , k] \rightarrow V$ maps an agent to a goal vertex. 
Time is discretized into time-steps. 
In every time-step, each agent occupies a single vertex and performs a single \emph{action}. 
%, i.e., if an agent executes action $a$ from vertex $v$ it will end up in vertex $v'$ in the next time-step. 
There are two types of actions: $wait$ and $move$. The result of a $wait$ action of an agent located in some vertex $v$ at some time-step 
is that the agent will stay at the same vertex $v$ at the next time-step. The result of a $move$ action is that the agent moves to an adjacent vertex $v'$ on the graph (i.e., $(v, v') \in E$).
A \emph{single-agent path} for agent $a_i$, denoted $\pi_i$, is a sequence of actions that start from $s_i$ and end in $g_i$. 
A solution to a MAPF is a set of single agent paths $\pi=\{\pi_1,\ldots,\pi_k\}$, one per agent, 
that do not \emph{conflict}. 
%We consider two types of conflicts: {\em vertex conflict} and {\em swapping conflict}.
Two single-agent paths have a {\em vertex conflict} if they aim to occupy the same vertex at the same time and a {\em swapping conflict} if they aim to traverse the same edge at the same time from opposing directions.  
The {\em sum-of-costs} (SOC) of a MAPF solution $\pi$ is the sum over the lengths of its constituent single-agent paths. 
It is often desirable to find minimal SOC solutions. 
%that have minimal SOC. 

% % Conflicts and a valid solution
% Let $\pi_i[x]$ denote the location of the agent after executing the first $x$ actions in $\pi_i$ starting from $s(i)$.
% A pair of single-agent plans $\pi_i$ and $\pi_j$ is said to have a \emph{vertex conflict} if 
% two agents occupy the same location at the same time, i.e., if there exists $x$ such that $\pi_i[x] = \pi_j[x]$. 
% Similarly, $\pi_i$ and $\pi_j$ are said to have a \emph{swapping conflict} if the agents swap locations at the same time, i.e. there exist $x$ such that $\pi_i[x+1] = \pi_j[x] \land \pi_i[x] = \pi_j[x+1]$. 
% Other types of conflicts have also been considered in the classical MAPF context~\cite{stern2019mapf} but in this work we consider  that a pair of single-agent plans have a conflict if they have either a vertex conflict or a swapping conflicts. 
% A solution to a MAPF is called \emph{valid} if its constituent single-agent plans do not have a conflict. The objective of classical MAPF algorithms is to find valid solutions.

% Some text on MAPF solvers
Some MAPF algorithms are \emph{complete and SOC-optimal}, i.e., guaranteed to return a valid optimal SOC solution if such exists.
%w.r.t. its SOC. 
 % Common solution cost functions are the sum of costs (SOC) and makespan, which are the sum and the maximum over the length of the MAPF solution's constituent single-agent plans. 
Primary examples of such algorithms 
%complete and optimal MAPF algorithms 
are CBS~\cite{CBS_2015}, ICTS~\cite{sharon2013increasing}, \astar+OD+ID~\cite{standley2010finding}, M*~\cite{wagner2011m}, BCP~\cite{BCP_2022}, and SAT-MDD~\cite{surynek2016efficient}. 
Other MAPF algorithms are complete but suboptimal. They are guaranteed to find a solution if such exists, but the cost of the returned solution may not be optimal. 
Examples of such algorithms are Push-and-Swap~\cite{luna2011push} and Kornhauser's algorithm~\cite{kornhauser1984pebble}. 
% In this work, we focus on incomplete MAPF algorithms, which are very common in real-world MAPF applications and are often much faster and can scale better than any complete algorithm~\cite{li2021anytime,leet2022shard}. We next present several incomplete algorithms. 

\paragraph{Prioritized Planning (PrP) and LNS2} 
PrP~\cite{bennewitz2001optimizing} is a simple yet very popular MAPF algorithm~\cite{leet2022shard,Varambally_2022,Zhang_2022,chan2023greedy}.  
% In addition, its running time is relatively small and it succeeds to find good solutions in many cases.  Indeed, PrP is still in use in practice and in recent suboptimal MAPF papers \cite{flatland_2021,ShardSystems_Li_Koenig_2022,Varambally_2022,Zhang_2022,chan2023greedy}. 
In PrP, the agents plan sequentially according to some predefined order. 
When the $i^{th}$ agent plans, it must avoid the paths chosen for all $i-1$ agents that have planned before it. 
PrP is agnostic to how a single-agent path is found for each agent as long as no conflicts with existing plans are generated. 
Such single-agent paths are found by a low-level search algorithm such as Temporal A* (\ta) or SIPPS, which are described below. PrP is simple and fast but might not be very effective in very dense environments due to possible deadlocks. Large Neighborhood Search for MAPF (LNS2)~\cite{li2022mapf} is an incomplete MAPF algorithm that aims to overcome some of the pitfalls of PrP. LNS2 starts by assigning paths to the agents even if they conflict. LNS2 then
applies a \emph{repair} procedure, where PrP is used to replan for a subset of agents, aiming to minimize conflicts with other agents. 
LNS2 repeats this repair procedure until the resulting solution is conflict-free. 

% LNS2 is more complicated than plain PrP but it was shown to outperform PrP in various circumstances.

% such that there are no conflicts between agents while minimize the number of collisions inside the group $A_s$. 

% repeats a repairing procedure until the entire solution becomes feasible. At each iteration of the repairing process,  This can be simply done by using PrP within the group and having each agent in its turn perform some low level algorithm. This repairing process is repeated until the resulting solution is conflict-free. LNS2 is more complicated than plain PrP but it was shown to outperform PrP in various circumstances.

%PrP is simple to implement and is widely used in practice due to its small runtime. 

% Notably, PrP is suboptimal and incomplete since its predeﬁned priority ordering can sometimes result in solutions of bad quality or even in a failure to ﬁnd a solution for solvable MAPF instances.

\paragraph{Temporal \astar, SIPP, and SIPPS}  Temporal \astar, SIPP, and SIPPS are algorithms that plan a path for a single agent under constraints. They are used in many MAPF algorithms, such as CBS~\cite{CBS_2015}, PBS~\cite{pbs_2019}, PrP, and LNS2. 
Temporal \astar\ (\ta) uses \astar\ on the {\em spatio-temporal state-space} in which a state is a vertex-time pair $(v,t)$
%where $v$ is a vertex $\in G$ and $t$ is a time-step. 
% We denote the \astar\ variant on this setting by \emph{Temporal \astar} (\ta). 
% A node in the \ta\ search space is a pair $(v,t)$ representing that the agent is planned to be in location $v$ at time $t$. 
\ta\ receives the start and goal locations of an agent, along with a list of vertex- and edge constraints. 
A vertex constraint is a pair $(v,t)$ specifying that the agent must not plan to be at $v$ at time-step $t$. An edge constraint $(e,t)$ specifies that the agent must not traverse edge $e$ at time $t$ in either direction.  \ta\ returns the shortest path that avoids the given constraints. 

Safe Interval Path-planning (SIPP)~\cite{phillips2011sipp} improves \ta\ by dividing time into intervals instead of times-teps to represent the time dimension. SIPP performs an A* search on a state space where each state is defined by a vertex and a safe (time) interval, representing that the agent occupies that vertex in this time interval and that this does not violate any given constraint. 
% These constraints are often derived from the paths of other agents that we would like to avoid, e.g., the agents that have already planned in PrP. \ta\ aims to find the optimal (the fastest) path to the goal location while satisfying the constraints. The temporal aspect also enables single-agent paths that include actions in which the agent waits in its place, which may be needed to satisfy the given constraints. 
Safe Interval Path-planning with Soft Constraints (SIPPS)~\cite{li2022mapf} generalizes SIPP to accept both \emph{hard} and \emph{soft} constraints. 
% A hard constraint is to avoid conflicts with the other agents already selected for replanning, and a soft constraint is to avoid conflicts with all other agents. \roni{Not clear, what is this replanning?} 
SIPPS returns a path that does not violate any hard constraints and minimizes the number of soft constraints violated. 
SIPPS is designed to run within LNS2. 
A node $n$ in the SIPPS search tree consists of four elements: a vertex $n.v$; a safe interval $[t_{start}^n, t_{end}^n)$, where $t_{start}^n$ is also called the earliest arrival time; an index $n.id$; and a boolean ﬂag $n.is\_goal$ indicating whether the node is a goal node.
The $f$-value of node $n$ is the sum of its $g$-value and $h$-value, where the $g$-value is set to $t_{start}^n$ and the $h$-value is a lower bound on the minimum travel time from vertex $n.v$ to vertex $g$. 
Each node $n$ also maintains a $c$-value, which is the (underestimated) number of soft collisions in the partial path from the root node to node $n$.
SIPPS first prioritizes nodes according to smaller $c(n)$ values. In case of a tie, it moves to the secondary priority and prefers nodes with smaller $f(n)=g(n)+h(n)$.

\paragraph{PIBT and LaCAM}

PIBT~\cite{okumura2022priority} is a MAPF algorithm that searches in the \emph{configuration space}.  
A configuration here is a vector representing the agents' locations at some time-step. 
PIBT searches in this space in a greedy and myopic manner, starting from the initial configuration of the agents and iteratively generating a configuration for the next time-step until reaching a configuration where all agents are at their goals. 
PIBT generates configurations recursively, moving every agent toward its goal while avoiding conflicts with previously planned agents. To avoid deadlocks, PIBT employs priority inheritance and backtracking techniques.  PIBT is very efficient computationally but is incomplete since it searches greedily in the configuration space. LaCAM~\cite{okumura2023lacam} also searches the configuration space in a similar manner. To ensure completeness, LaCAM adds constraints to the configuration generation process to ensure that it can eventually reach every possible configuration. LaCAM$^*$~\cite{okumura2023lacam} builds on LaCAM but improves its configuration generation process. LaCAM$^*$ may also continue the search in an anytime fashion, and it eventually converges to the optimal solution.

\subsection{Lifelong MAPF (LMAPF)}
\label{sec:lifelong}

{\em Lifelong MAPF (LMAPF)}~\cite{li2021anytime} is a version of {\em online MAPF}~\cite{vsvancara2019online} where agents continuously receive new tasks from a task assigner  (outside the path-planning system). When an agent reaches its current goal, it receives a new goal to travel to from the task assigned. 
The task in classical MAPF is to minimize the cost of the solution. 
Since solving LMAPF involves solving multiple MAPF problems, the efficiency of LMAPF algorithms is commonly measured by the overall system {\em throughput} achieved, which is measured by the number of tasks completed in a given period of time~\cite{RHCR_Li_Koenig_2021,fail_policies_2023}. 

The Rolling-Horizon Collision Resolution framework (RHCR) \cite{RHCR_Li_Koenig_2021}, solves MAPF problems by repeatedly planning the next $k$ steps ($k$ is the {\em horizon} parameter) to execute based on the current locations and goals of the agents. The agents then move $w \leq k$ steps, where $w$ is a {\em window} parameter, and the process repeats. RHCR is particularly useful in LMAPF as new goals are generated on the fly and there is no point in planning long paths. 
% \arseni{Roni, is it safe to say that RHCR is a standard also for classical MAPF?..}
% \ariel{Yes, Roni please answer}
% \roni{No, I'd say RHCR is the standard for LMAPF. In general it was designed for LMAPF}

% LMAPF and RHCR

\paragraph{Congestion avoidance techniques}
Several approaches were proposed to avoid the emergence of congested areas in LMAPF.
Han et al.~\cite{han2020ddm} used database heuristics that are customized for every map separately.
They also proposed a Space Utilization Optimization heuristic for clever tie-breaking that still preserves optimality~\cite{han2022optimizing}. 
Surynek et al. ~\cite{skrynnik2024learn} developed a reinforcement learning (RL) approach for a partially observable variant of MAPF. 
In our work, we neither restrict observability nor do we 
require a pre-processing stage for training. 
% That being said, all of them are orthogonal to our work and potentially can improve the performance when combined.
Therefore, we do not compare our work with these approaches.
Chen et al. ~\cite{chen2024traffic} proposed to exploit time-independent routes to compose a heuristic for the PIBT algorithm that predicts future expected congestion. 
Our method for incorporating APFs in PIBT is inspired by their work.

\subsection{Artificial Potential Fields}
% \roni{Maybe we want such a subsection that mentions what are APFs and existing applications of APFs}
APFs are commonly used for single-agent motion planning in continuous spaces \cite{Barraquand_1992,Fox_1997,Yunfeng_2000,daily2008harmonic,mac2016heuristic,zhang2018path,shin2021hybrid}.
Vadakkepat et al. \cite{Vadakkepat_2000} studied continuous spaces with moving obstacles.
Hagelb{\"a}ck et al. \cite{hagelback2008using,hagelback2009multi} used APFs in real-time strategy games to avoid obstacles.
Other works dealt with multiple agents. Liu et al.~\cite{liu2017formation}
used APFs in a problem of formation control; Dinh et al.~\cite{dinh2016multi} introduced Delegate MAS that simulated food foraging behavior in ant colonies; and many works \cite{semnani2020multi,fan2020distributed,agrawal2022dc,dergachev2021distributed} introduced reinforcement learning algorithms (RL) that incorporated AFPs together with Optimal Reciprocal Collision Avoidance (ORCA) \cite{Berg_2011_orca} or Force-based motion planning (FMP) \cite{semnani2020multi}.
Some algorithms include APFs as part of the environment~\cite{bettini2022vmas}. 
To the best of our knowledge, we are the first to apply APFs to solve MAPF problems.

% All MAPF algorithms use TA. Explain how. 

% We modify TA to use PF

%%%%%%%%%%%%%%%%%%%%%%%%%%%%%%%%%%%%%%
%%%%%%%%%%%%%%%%%%%%%%%%%%%%%%%%%%%%%%
%%%%%%%%%%%%%%%%%%%%%%%%%%%%%%%%%%%%%%
%%%%%%%%%%%%%%%%%%%%%%%%%%%%%%%%%%%%%%
%%%%%%%%%%%%%%%%%%%%%%%%%%%%%%%%%%%%%%
%%%%%%%%%%%%%%%%%%%%%%%%%%%%%%%%%%%%%%
%%%%%%%%%%%%%%%%%%%%%%%%%%%%%%%%%%%%%%
%%%%%%%%%%%%%%%%%%%%%%%%%%%%%%%%%%%%%%
%%%%%%%%%%%%%%%%%%%%%%%%%%%%%%%%%%%%%%
%%%%%%%%%%%%%%%%%%%%%%%%%%%%%%%%%%%%%%
%%%%%%%%%%%%%%%%%%%%%%%%%%%%%%%%%%%%%%
\section{Direct Artificial Potential Fields}
\label{sec:dapf}
% \roni{Here should come a place where Arseni explains how APFs can be used to solve MAPF.}
In this section, we present a {\em direct application} of APFs (DAPF) to solve MAPF problems.
In DAPF, at every time-step each agent $a_i$ sums a set of repulsion and attraction ``forces'' and moves in the corresponding direction. 
These ``forces'' consist of repulsion from the locations of all other agents and attraction to the location of the goal of $a_i$.  
We experimented with different functions for these repulsion and attraction ``forces'' in our MAPF context. 
The following functions worked reasonably well in our experiments. 

\paragraph{Repulsion forces:} for every agent $a_i$, we create a repulsion function $APF_i$ 
based on its current location $v_i$. 
\begin{equation}
    APF_i(v)=
\begin{cases} 
      0 & \text{if } d(v_i,v) > \maxdist \\
      w\cdot\shape^{-d(v_i,v)}     
      & \text{otherwise} 
\end{cases}
\label{eq:direct_apf}
\end{equation}
where $\maxdist$, $\gamma$, and $w$ are predefined parameters and  $d(v_i,v)$ is the minimal distance between $v$ and $v_i$.\footnote{This could be the exact minimal distance in the graph or some estimation heuristic on it. In our experiments, every agent location is associated with a cell in a grid. 
We used Manhattan distance.}
The $w$ parameter controls the strength of the repulsion, $\gamma$ controls the rate of decay, i.e., how fast its intensity of the repulsion decreases while moving away from its source, and $d_{max}$ defines how far away from $v_i$ the repulsion affects the cost.\footnote{We illustrate this in the supplementary material.}

\paragraph{Attraction forces}
For every agent $a_i$, we create an attraction function $GoalAPF_i$, based on its goal $g_i$:
\begin{equation}
    GoalAPF_i(v)=
\begin{cases} 
       h(g_i,v)       & \text{for agent $a_i$} \\
      0 & \text{for other agents} \\
      
\end{cases}
\end{equation}
where $h(g_i,v)$ is a precalculated heuristic estimation on the distance from vertex $v$ to $g_i$. 
%\arseni{Can we add something like this: Note that the difference in cost of $GoalAPF_i(v)$ between neighboring locations is at most $1$ so that the scale is comparable with $APF_i(v)$.}
Given these repulsion and attraction functions, DAPF moves in every time-step 
each agent $a_i$ located at $v_i$ to the adjacent location $v'_i$ that minimizes 
\begin{equation}
Total\_APF(v'_i)=\sum_{j\neq i} APF_j(v'_i)+\sum_j GoalAPF_i(v'_i)
\end{equation}
Ties are broken randomly. 
Collisions are avoided in a prioritized planning manner. The agents plan sequentially (with some random order), where every agent cannot occupy the current locations of other agents and the locations reserved by previously planned agents.

% \footnote{We note again that we aim to minimize  $Total\_APF(v'_i)$. Indeed, at the goal the $GoalAPF_i(goal)=0$ and if the goal is located far away from the other agents then their repulsion forces are also very small or even zero (depending on the parameters).}
% as follows. 
% Assume $a_i$ is located at $v_i$ and let $l_j$ be a location adjacent to $v_i$.  For every $l_j$ agent $a_i$ sums all of the potential fields that affect $l_j$. Then, $a_i$ moves to the adjacent location with the lowest value (ties are broken randomly). With these AFPs, agents push away each other reducing potential conflicts. In terms of runtime complexity of DAPF, the computation for each agent $a_i$ is linear in the number of other agents. 

 \begin{figure} [t]
 \begin{center}
    \subfloat[]{\includegraphics[scale=0.27]{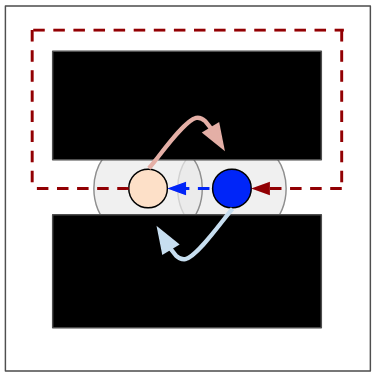}}
    \hspace{10pt}
    \subfloat[]{\includegraphics[scale=0.27]{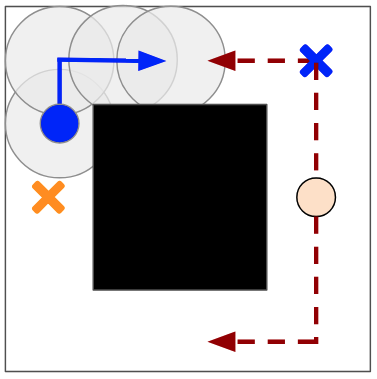}}
 \caption{(a) An instance that cannot be solved by DAPF. The dotted lines depict a PrP solution. (b) Two agents solve LMAPF. The $X$ shapes represent goal locations. A solid blue line shows the direction of a chosen path for the blue agent. Dashed red lines represent alternative $k$-length paths for the orange agent. The agent will prefer the bottom path because of the APFs of the path of the blue agent.}
 \label{Fig:pure_pfs_example}
 % \vspace{-18pt}
 \end{center}
 \end{figure}

DAPF is very efficient computationally. In every iteration, each agent incurs a runtime of $O(k\cdot (d_{max})^2)$, where 
% $d_{max}^2$ is the number of next possible locations and 
$k$ is the number of agents and $d_{max}$ is the radius of influence of APFs. 
So, the overall runtime complexity of DAPF is $O(k^2\cdot (d_{max})^2\cdot N)$, where $N$ is the number of iterations.
% DAPF is very efficient computationally. In every iteration, each agent incurs a runtime of $O(d\cdot k)$, where $d$ is the number of next possible locations and $k$ is the number of agents. 
% The overall runtime complexity of DAPF is $O(d\cdot k^2\cdot N)$, where $N$ is the number of iterations.
% DAPF can also be distributed by having each agent compute its APFs and corresponding actions independently and concurrently. In this case, each agent computes its APFs and corresponding actions independently and concurrently, thus the overall runtime complexity becomes $O(d\cdot k\cdot N + d\cdot k)$, where the last component $d\cdot k$ is for the location selection process. This is similar to several other MAPF algorithms~\cite{Velagapudi_2010}.
% \arseni{In the paper: $k$ - for number of agents, $k'$ - for a subset of $k$ agents, and $n$ - for node. I changed back all the paper according to those notations. The $n$ notation is better for node, I think.}

In spaces with relatively few obstacles and a small number of agents, using DAFP is very fast. 
However, it performed poorly in our experiments when the environment became more dense. 
This is because there are cases where long-term planning is needed and a na\"{i}ve usage of APFs, as done by DAPF, is insufficient no matter which parameters we choose. For example, consider the MAPF problem in Figure \ref{Fig:pure_pfs_example} (a). Solid lines point to the agents' goal locations. Here, DAPF fails to find a solution.
Agents are pushed into and pulled back to the middle corridor, preventing agents from swapping positions. The possible solution is for the orange agent to bypass the obstacle and reach its goal (dashed line in Figure \ref{Fig:pure_pfs_example} (a)). 
In the following sections, we propose a range of techniques for using APFs within existing MAPF algorithms.

% \ariel{Explain why? Agents are pushed from and pulled into the same location??}

%  \begin{figure} [!ht]
% %\vspace{-8pt}
%    % \hspace{2pt}
%  \begin{center}
%   \includegraphics[scale=0.15]{pics/pure_pfs_example.png}
% % \vspace{-10pt}
%  \caption{APFs will not solve the instance.}
%  \label{Fig:pure_pfs_example}
%  % \vspace{-18pt}
%  \end{center}
%  \end{figure}

%%%%%%%%%%%%%%%%%%%%%%%%%%%%%%%%%%%%%%
%%%%%%%%%%%%%%%%%%%%%%%%%%%%%%%%%%%%%%
%%%%%%%%%%%%%%%%%%%%%%%%%%%%%%%%%%%%%%
%%%%%%%%%%%%%%%%%%%%%%%%%%%%%%%%%%%%%%
%%%%%%%%%%%%%%%%%%%%%%%%%%%%%%%%%%%%%%
%%%%%%%%%%%%%%%%%%%%%%%%%%%%%%%%%%%%%%
%%%%%%%%%%%%%%%%%%%%%%%%%%%%%%%%%%%%%%
%%%%%%%%%%%%%%%%%%%%%%%%%%%%%%%%%%%%%%
%%%%%%%%%%%%%%%%%%%%%%%%%%%%%%%%%%%%%%
%%%%%%%%%%%%%%%%%%%%%%%%%%%%%%%%%%%%%%
%%%%%%%%%%%%%%%%%%%%%%%%%%%%%%%%%%%%%%
\section{Integrating APFs in TA* and SIPPS}
% \section{Temporal \astar\ with APFs}
\label{pf_descr}

% Intro: motivation - what TAPF aims to achieve
% This section proposes a technique that uses APFs within existing MAPF algorithms. 
Many MAPF algorithms internally use either \ta\ or SIPPS to find paths for individual agents. We propose to use APFs 
%as a way to impose additional {\em soft constraints} 
within these single-agent path-finding algorithms such that the resulting path not only avoids collisions with the paths of other agents but also attempts to keep distance from them by taking into consideration the repulsion of their APFs.  We refer to our \ta\ and SIPPS variants as \emph{Temporal \astar\ with APFs} (\tapf) and \emph{SIPPS with APFs} (\sippsapf), respectively.   

% ORIGINAL: in \ta\ such that the resulting path not only avoids collisions with the paths of other agents (which is a hard constraint) but also attempts to keep distance from them by considering the repulsion of their APFs.  We refer to our \ta\ variant as \emph{Temporal \astar\ with Artificial Potential Fields} (\tapf).   

% TA*+PF high-level concept
\subsection{Temporal \astar\ with APFs}
\tapf\ is a single-agent path-finding algorithm. It accepts as input
a tuple $\langle G(V,E),s,g,\{\pi_1,\ldots,\pi_{k'}\} \rangle$ 
where $G$ is the graph of possible locations the agent can occupy ($V$) and the allowed transitions between them ($E$); 
$s$ and $g$ are the start and goal locations of the path planning agent; 
and $\pi_i$ is a path for agent $a_i$ for every $i=1,\ldots,k'$. 
The output of \tapf\ is a path from $s$ to $g$ that does not conflict with any of the paths $\pi_1,\ldots,\pi_{k'}$. 
The given set of paths $\{\pi_1,\ldots,\pi_{k'}\}$ depend on the particular MAPF algorithm in use. 
For example, in PrP \tapf\ will be given the paths planned for the higher-priority agents. 
In LNS2, the given set of paths includes the paths already planned for the other agents within the neighborhood of the planning agent.

% The inputs of \tapf\ is a tuple $\tuple{G(V,E),s,g,\{\pi_1,\ldots,\pi_n\}}$ 
% where $G$ is the graph of possible locations ($V$) that the agent can occupy and allowed transitions between them ($E$); 
% $s$ and $g$ are the start and goal locations of agent $a_i$, respectively; 
% and $\pi_j$ is a path for agent $a_j$. 
% The output 
% % Let $N$ be the set of agents that agent $a_i$ wishes to avoid and let $\pi_j$ be the . 
% For every agent $a_j \in N$ (i.e., for $j=1,\ldots, k$) $\pi_j$ is the path of an agent $a_j$.
% For every agent $a_j \in N$ we create a potential field $APF_j$, which is a function that maps every location-time pair $(v,t)$ to a real number representing the added penalty incurred by $a_j$ for planning to occupy $v$ at time $t$.
% %\ariel{Not clear. you have $V$ locations but how many time-steps? Or do we do this when planning for a given agent for each possible move? Also who are the $n$ other agents? Say something about them} 
% \tapf\ considers these penalties when computing the cost of every move. 

% How to define PF_i
To bias the resulting path to keep distance from these paths, \tapf\ creates for every path $\pi_i\in\{\pi_1,\ldots,\pi_{k'}\}$ a repulsion APF function $APF_i$ that maps every location-time pair $(v,t)$ to a real number representing the added penalty incurred by planning to occupy $v$ at time $t$.
%\ariel{Not clear. you have $V$ locations but how many time-steps? Or do we do this when planning for a given agent for each possible move? Also who are the $n$ other agents? Say something about them} 
\tapf\ considers these penalties when computing the cost of every move. 
There are multiple ways to define these repulsion APF functions and aggregate them into a single penalty cost. 
We experimented with several options and have found the following implementation to work best in our experimental setup. 
Similar to DAPF, the APF induced by agent $a_i$ on location $v$ and time $t$ is computed as follows:
\begin{equation}\label{eq:afp}
    APF_i(v,t)=
\begin{cases} 
      0 & \text{if } d(v,\pi_i[t]) \geq \maxdist \\
      w\cdot\shape^{-d(v,\pi_i[t])}    
      & \text{otherwise} 
\end{cases}
\end{equation}
where $\maxdist$, $\gamma$, and $w$ are predefined parameters and $d(v,t,\pi_i[t])$ is the minimal distance between $v$ and $\pi_i[t]$.\footnote{Again, we used Manhattan distance.} 
% \footnote{In our experiments on grids we computed distance as the floor of the Euclidean distance between these grid cells, i.e., $\lfloor||v-\pi_i[t]||_2\rfloor$.} 
The purpose of each parameter is as described in Section~\ref{sec:dapf}. % and is demonstrated in Figure \ref{fig:apfs_params}.
The only difference between this computation method compared to the one used by DAPF (Eq.~\ref{eq:dapf}) is the introduction of the time dimension.
Hence, in each time-step $t'$ an agent considers only APFs that are calculated at $t'$.
% Thus, in each time-step the agent may be influenced only by APFs that are defined exactly for the same time-step.
% Intuitively, 
% (1) \maxdist, defines how far the APF of an agent will influence the cost;
% (2) $\shape$, defines how fast the intensity of an APF declines while moving away from its source;
% and 
% (3) $w$, defines the importance of the APFs w.r.t. the actual path cost. 
In the special case of $\maxdist=0$,  $APF_i(v,t)$ is always 0. 
This corresponds to plain \ta, which only considers conflicts where agents are exactly at the same location (distance 0 from each other). 
To aggregate all the APFs we used a simple sum. 
% That is, the cost of moving into location $v$ at time $t$ while considering the APFs is
That is, the APF cost of moving into location $v$ at time is
\begin{equation}
    cost_{APF}(v,t)=\sum_{i \in \{1,\ldots,k'\}}  APF_i(v,t)
\end{equation}

% \arseni{The computation can be done lazily.}

% \arseni{a little repetition from the prev section - is it fine? AF: indeed repetitions. }

% \roni{Add here a discussion and maybe an example of where should the cost be added: to g or to h}
We considered incorporating our APF-inspired cost function within \ta\ in two places: as part of the heuristic evaluation function or as part of the edge cost function. 
For a node $n$ generated by \ta, the former adds a penalty to $h(n)$ and the latter adds a penalty to $g(n)$. 
We initially incorporated our APF-inspired cost function in $h(n)$ and observed poor results. To understand these results, consider what each value ($h$ and $g$) represents and its objective. 
$h(n)$ represents an estimate of the cost from $n$ to the goal. It is designed to guide the search towards finding the goal faster. 
In contrast, $g(n)$ represents the cost of the best path found so far from the start to $n$. \astar, and subsequently \ta, are designed to minimize the edge cost function and correspondingly find the path to the goal with the lowest $g$-value. Thus, incorporating APFs into heuristic functions does not directly affect what path \ta\ returns but rather how quickly the search finds it. In contrast, incorporating APFs into the edge cost function (and thus in the $g$ value) directly results in \ta\ returning paths that optimize for avoiding other agents' paths.
 \begin{figure} [!ht]
%\vspace{-8pt}
   % \hspace{2pt}
 \begin{center}
  \includegraphics[scale=0.22]{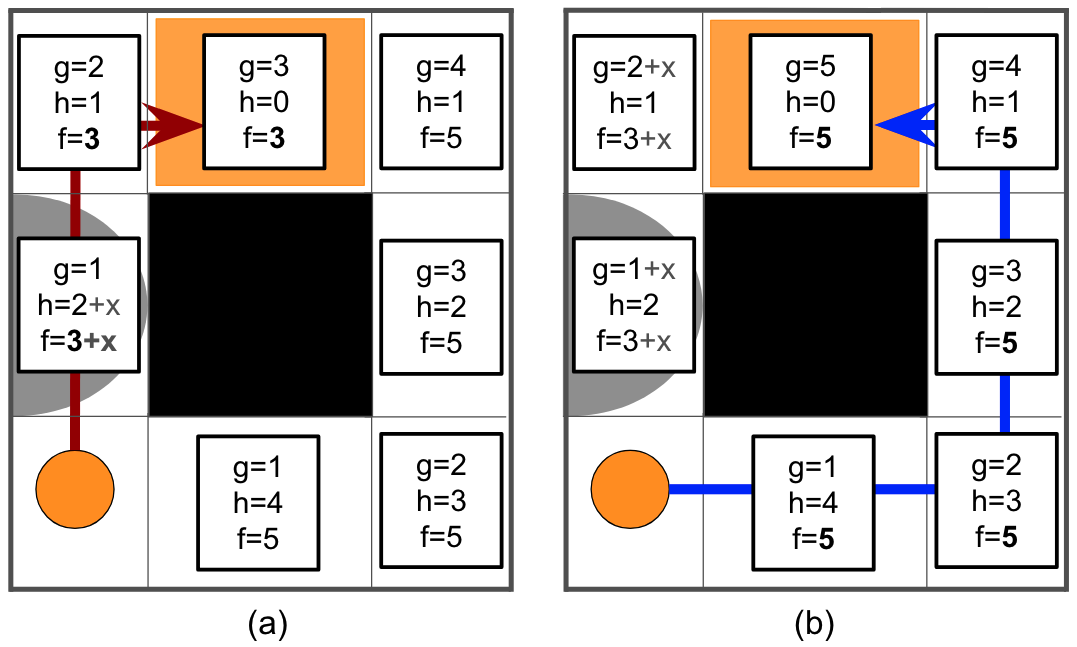}
% \vspace{-10pt}
 \caption{The orange circle and the orange square are the agent's start and goal locations, respectively. $x$ represents the cost of APFs. 
 \astar\ is executed. $g, h$ and $f$ are the components of \astar\ nodes. (a) With $x\leq2$, an agent always picks a red path wherever $x$ is added. (b) With $x>2$, the agent picks a blue path if $x$ is added to $g$. Otherwise, if $x$ is added to $h$, it continues to choose a red path nonetheless.
 % Lines represent possible paths. A gray area shows APFs of another agent that adds to the cost of a location. Numbers represent the AFPs costs added to $g$ or to $h$ components of \astar.
 }
 \label{Fig:g_or_h}
 % \vspace{-18pt}
 \end{center}
 \end{figure}

Consider Figure \ref{Fig:g_or_h}, where an agent needs to go from the orange circle to the orange square. 
The components $g$, $h$, and $f$ of an \astar\ algorithm are presented inside the locations. The gray zone shows the APFs of another agent that adds to the cost $x$ of moving through the location in the middle left of the figure. 
Consider the case where $x > 2$.
A red line is an optimal path that \astar\ would choose if the costs were added to a $h$ component (Fig. \ref{Fig:g_or_h} (a)). 
A blue line is an optimal path if the cost is part of a $g$ component, and this is the variant that we want to implement (Fig. \ref{Fig:g_or_h} (b)) because we want to avoid the congested gray cell. In other words, $h$ is not being aggregated along paths. It is only defined for individual locations and only attracts the agent towards the goal. In contrast, $g$ is being aggregated and an edge with a large weight will be carried over to all its descendants, and this is what we want. 

% \roni{Good to have here an example, showing that APFs in the h value won't help enough}\roni{@Ariel,@Roie,@Arseni: is the paragraph above Ok? }

Therefore, in \tapf\ we add our APF-inspired cost function to the $g$ value of a search node when a node is generated, as follows:
\begin{equation}
 g_{APF}(n) = g_{APF}(p) + cost(p, n) + cost_{APF}(v,t)     
\end{equation}
\noindent where $p$ is the parent of $n$, $cost(p,n)$ is the cost of moving the agent from $p$ to $n$, and $(v,t)$ is the vertex and time-step of node $n$. \tapf\ runs \ta\ and prioritizes nodes for expansion according to
\begin{equation}
 f(n)=g_{APF}(n)+h(n)    
\end{equation}

% \roni{For consistency, changed here also}

Figure~\ref{Fig:pure_pfs_example}(b) illustrates an example case, where APFs help to pick a better path out of two options. In this example, two agents work in a RHCR framework. The blue agent plans first and goes directly to its goal. An orange agent has two alternative paths around the obstacle. However, the APFs of the blue agent (gray circles) will cause the orange agent to prefer the bottom path and therefore avoid future conflicts.

Using APFs incurs some computational overhead, compared to vanilla \ta. 
This overhead is due to the need to compute APFs for every newly generated path. 
The computational complexity of generating an APF for the path of a single agent is $O((d_{max})^2\cdot l)$, where  $(d_{max})^2$ is the maximal number of nodes affected by the APF induced by an agent occupying a single vertex at a specific time step; and $l$ is the length of the longest path. 
This computation is done for each of the $k$ agents, adding a total overhead of $O(k\cdot (d_{max})^2\cdot l)$. 
\subsection{SIPPS with APFs}
\label{apf_sipps}

% Intro: motivation - what \sippsapf aims to achieve

%A more recent low-level solver in multiple MAPF algorithms is SIPPS~\cite{li2022mapf}.
%As was described in the background section,
\sippsapf is based on SIPPS. 
Thus, a search nodes $n$ in \sippsapf represents a vertex $n.v$ and a time interval $[t_{start}^n, t_{end}^n)$, and the input to \sippsapf includes a set of \emph{soft constraints} it aims to minimize violating.  

% How to define APF_i
We define an APF cost for a \sippsapf node $n$ as follows:
\begin{equation}
\label{eq:cost_sipps}
    cost_{APF}(n)=\max_{t\in[t_{start}^n, t_{end}^n)}{\sum_{i \in \{1,\ldots,k'\}}  APF_i(v,t)}
\end{equation}
where $APF_i(v,t)$ is defined in \tapf . 
In other words, $cost_{APF}(n)$ represents the highest APFs an agent can encounter during its time interval.

% \sippsapf high-level concept 
In SIPPS, the open list is sorted first according to 
the number of soft collisions violated and then according to $f(n)$. 
\sippsapf follows the same sorting rule, but add the APF costs in each of these components. 
% two components, first $c(n)$ and then $f(n)$. 
% $c(n)$ is the number of soft collisions violated, i.e., constraints caused by a collision with another path, and $f(n)=g(n)+h(n)$ is the standard \astar\ evaluation function. 
We explored adding the \sippsapf APF costs separately in each of these components, but this yielded limited gains.  
% We first tried to add APFs to $c(n)$ only, and the effect was modest. Then, we tried to add APFs only to the $g(n)$ component as in \tapf\, and the improvement was also small.
% This yielded limited gains. However, incorporating APF costs in both components resulted in a significant performance boost. 
% This was done as follows. 

%Hence, the final formal definition is presented as follows.

In more details, \sippsapf maintains for each node $n$
two additional values, $c_{APF}(n)$ and $g_{APF}(n)$, 
which are computed as follows when $n$ is generated: 
\begin{align}
   c_{APF}(n) = & c_{APF}(p) + c(p, n) + cost_{APF}(n) \\
   g_{APF}(n) = & t_{start}^n + cost_{APF}(n) 
\end{align}
where $p$ the parent of $n$, $c(p, n)$ are the number of soft constraints violated when moving from $p$ to $n$. 
\sippsapf sorts open list first based on $c_{APF}$ and then based on $g_{APF}+h(n)$. 
% \roni{Worth re-reading the above paragraphs}

% and sorts its open list first accordingly to  
% $c_{APF}$ and then according to $g_{APF}(n)+h(n)$. 
% These values are 

% \sippsapf first sorts the open list according to the following function: 
% \begin{equation}
%    c_{APF}(n) = c_{APF}(parent) + c(n) + cost_{APF}(n)     
% \end{equation}
% As a secondary sort function, \sippsapf uses $g_{AFP}(n)+h(n)$, where: 
% \begin{equation}
%     g_{APF}(n) = t_{start}^n + cost_{APF}(n)     
% \end{equation}
% \roni{IMPORTANT: in APF, the APF costs were accumulated, here they are not (see the differences between the definition of TAPF. Shouldn't it be: 
%     \begin{equation}
%        c_{APF}(n) = c_{APF}(parent)+ c(n) + cost_{APF}(n)     
%     \end{equation}
%     PLEASE THINK ABOUT THIS CAREFULLY DO NOT JUST WRITE IT, I AM NOT SURE THIS IS CORRECT IN C(N)!
%     As a secondary sort function, \sippsapf uses: 
%     \begin{equation}
%         g_{APF}(n) = g_{APF}(parent) + cost(parent, n) + cost_{APF}(n)     
%     \end{equation}
% }
% \arseni{[Arseni: The $c$ values are accumulated. I just checked the SIPPS paper and our code. Regarding the $g_{APF}(n)$ part, the formula is correct. In SIPPS, the algorithm works in time intervals, so the lower bound of every SIPPS node is set to be $t_{start}^n $, which is the start of the time interval. To this value, we have added the APF costs. As far as I remember, we indeed tried a lot of things there, including the accumulation of $g$ values. Did not work well. 
% I have corrected the text above a bit.]}

The analysis of the runtime complexity overhead incurred by APFs in \sippsapf\ is similar to the analysis described for \tapf. An additional computational cost is incurred due to the maximization of the time steps in the relevant safe interval (Eq. \ref{eq:cost_sipps}). 
So, the overhead equals $O(k\cdot d_{max}^2\cdot l^2)$, where $k$, $d_{max}$, and $l$ are defined as earlier. 
% the number of agents, the radius of APF influence, and the length of the longest path, respectively. 
% The runtime complexity overhead incurred by APFs is similar to one of \tapf\ with an additional cost of a $max$ computation (Eq. \ref{eq:cost_sipps}).
% So, the overhead equals $O(|V|\cdot k\cdot l^2)$, where $|V|$, $k$, and $l$ are the number of nodes, the number of agents, and the length of the longest path, respectively. 

\subsection{Completeness of TA*+APF and SIPPS+APF} 

It is important to note that our new idea of using APFs inside the $g$-value (denoted above as $g_{APF}$) is only considered when computing the priority of nodes in the open list. However, every node in TA*+APF and in SIPPS+APF still maintains the original spatio-temporal information as in regular TA* and SIPPS. In particular, the original $g$-value, which adds up the physical steps is still maintained to determine the exact time step for identifying conflicts as well as to prune duplicates.
Thus, the search spaces are the same with and without APFs, and the only difference is the order in which nodes are expanded. 
Consequently, completeness is preserved for both algorithms.

\section{PIBT and LaCAM with APFs}
\label{apf_pibt}

% PIBT~\cite{okumura2022priority} and LACM~\cite{okumura2023lacam} are state-of-the-art MAPF algorithms that search in a \emph{configuration space}.  
% A configuration here is a vector of vertices, one per agent, representing the agents' locations in some time-step. 

% Each configuration is created recursively by moving every agent toward its goal while avoiding conflicts with previously planned agents, and using
% priority inheritance and backtracking to avoid deadlocks. 
Both PIBT~\cite{okumura2022priority} and LaCAM~\cite{okumura2023lacam} use a heuristic to prioritize the actions of agents when generating configurations. In these algorithms, each agent chooses its next action by sorting the vertices adjacent to it based on a heuristic estimate of their distance from the goal. 
We propose to add APFs to these heuristic estimates and refer to our PIBT variant as \emph{PIBT with APFs} (PIBT+APF).
Specifically, when agent $i$ in PIBT+APF picks the next location, it continues to calculate several steps forward, defined by the $t_{max}$ parameter.
Those steps follow agent $i$'s optimal path to its goal and ignore other agents. 
Then, PIBT+APF builds APFs around these steps with Equation \ref{eq:afp} and sums up all the values in the time dimension. 
\begin{equation}
    PIBT\_APF_{i}(v)=\sum_{j \in \{t_{curr},\ldots,t_{curr} + t_{max}\}}  APF_i(v, j)
    \label{eq:pibt_apf}
\end{equation}
where, $t_{curr}$ defines the current time step. 
As a result, an APF is created for an agent $i$.
Finally, when every agent sorts the vertices adjacent to its current location, it considers the APFs created by the previously planned agents and sums all the APFs for every neighboring vertex as follows:
\begin{equation}
    cost_{APF}(v)=\sum_{i \in \{1,\ldots,k'-1\}}  PIBT\_APF_{i}(v)
    \label{eq:dapf}
\end{equation}
\noindent where $PIBT\_APF_{i}(v)$ is defined above (Eq.~\ref{eq:pibt_apf}).
Finally, we sort the vertices according to $h(v) + cost_{APF}(v)$. 
The overhead incurred by PIBT\_APFs in terms of run-time complexity is $O(k \cdot d_{max}^2 \cdot t_{max})$ per time-step, corresponding to computing $cost_{APF}$ for all $k$ agents and nodes in radius $d_{max}$. The total overhead is  $O(k\cdot d_{max}^2 \cdot t_{max} \cdot l)$, where $l$ is the length of the longest path.
Our APF-enhanced version of the LaCAM algorithm uses PIBT+APF for its low-level search.
This approach of adapting APFs to PIBT, in part, is inspired by the work of Chen et al. \cite{chen2024traffic} in which the authors use congestion-avoiding paths to navigate agents. 
Our work can be viewed as the generalization of this approach, where we consider not only the future steps of agents, but also the surrounding locations of these steps. We use their method as a baseline in our experiments.
% \ariel{you should say a sentence or two about how you added APF to LACAM}
% Specifically, when sorting the vertices adjacent to the $k'$-th agent, we consider the actions chosen by the previously planned agents $(1, \dots, k'-1)$ with Equation \ref{eq:direct_apf}. 
% Then, we sum all the APFs for every neighboring vertex as follows:
% \begin{equation}
%     cost_{APF}(v)=\sum_{i \in \{1,\ldots,k'-1\}}  APF_i(v)
%     \label{eq:dapf}
% \end{equation}

% This use of APFs is reminiscent of the DAPF algorithm presented above and may suffer from similar limitations.  
% The overhead incurred by APFs in terms of runtime complexity is $O(|V|\cdot k)$ per time-step, corresponding to computing $cost_{APF}$ for all $|V|$ nodes and $k$ agents. Therefore, the total overhead is  $O(|V|\cdot k\cdot l)$, where $l$ is the length of the longest path.

% \ariel{Major issue: Do we have cases where DAPFs do manage to succeed in some way. Do we want to provide experimental evidence to these and show the advantage here too? That is, in such cases DAPFs are very fast to compute and are successful.}

%%%%%%%%%%%%%%%%%%%%%%%%%%%%%%%%%%%%%%
%%%%%%%%%%%%%%%%%%%%%%%%%%%%%%%%%%%%%%
%%%%%%%%%%%%%%%%%%%%%%%%%%%%%%%%%%%%%%
%%%%%%%%%%%%%%%%%%%%%%%%%%%%%%%%%%%%%%
%%%%%%%%%%%%%%%%%%%%%%%%%%%%%%%%%%%%%%
%%%%%%%%%%%%%%%%%%%%%%%%%%%%%%%%%%%%%%
%%%%%%%%%%%%%%%%%%%%%%%%%%%%%%%%%%%%%%
%%%%%%%%%%%%%%%%%%%%%%%%%%%%%%%%%%%%%%
%%%%%%%%%%%%%%%%%%%%%%%%%%%%%%%%%%%%%%
%%%%%%%%%%%%%%%%%%%%%%%%%%%%%%%%%%%%%%
%%%%%%%%%%%%%%%%%%%%%%%%%%%%%%%%%%%%%%
\section{Experimental Study}

We conducted an experimental evaluation comparing the use of APFs within PrP~\cite{prp_2015}, LNS2~\cite{li2022mapf}, PIBT~\cite{okumura2022priority}, LaCAM~\cite{okumura2023lacam}, and LaCAM$^*$~\cite{okumura2023lacam}, where PrP and LNS2 are implemented twice: once with \ta\ and once with SIPPS as their low-level solver. 
With APF-enhanced versions, this adds up to a total of 14 algorithms.
Versions that use APFs are denoted in our plots by dashed lines. 
As an additional baseline, we implemented \emph{PIBT with Guide Path} (L-PIBT+GP) from Chen et al. \cite{chen2024traffic}. 
% We explored different heuristics for ordering the agents, including path length, amount of neighbors, and expected finishing times. 
% Our experiments with these heuristics showed no significant advantage over random ordering, hence they are not presented in the paper.
% Benchmark
All experiments were carried out on four different maps from the MAPF benchmark~\cite{stern2019mapf}: \emph{empty-32-32}, \emph{random-32-32-10}, \emph{random-32-32-20} and \emph{room-32-32-4}, as they present different levels of difficulty. 
% The maps are visualized in Figure~\ref{fig:rsoc}. 
% \begin{figure} [ht]
%  \begin{center}
%   \includegraphics[scale=0.5]{pics/empty-32-32_small.png}
%   \includegraphics[scale=0.5]{pics/random-32-32-10_small.png}
%   \includegraphics[scale=0.5]{pics/random-32-32-20_small.png}
%   \includegraphics[scale=0.5]{pics/room-32-32-4_small.png}
% % \vspace{-10pt}
%  \caption{MAPF Grids}
%  \label{fig:maps}
%  \end{center}
%  \end{figure}
The number of agents used in our experiments ranged from 50 to 450.  We executed 15 random instances per every number of agents, map, and algorithm.
The APF parameters used were
$w=1$, $\maxdist=4$, and $\shape=2$ for \tapf, 
$w=0.1$, $\maxdist=3$, and $\shape=3$ for \sippsapf, and 
$w=0.1$, $\maxdist=2$, $\shape=3$, and $t_{max}=2$ for PIBT+APF, which were observed to work best in general across all grids.\footnote{A sensitivity analysis of the impact of these parameters is discussed later and in the supplementary material.}
% Unless specified otherwise, the APF parameters used were $w=1$, $\maxdist=4$, and $\shape=2$, which were observed to work best in general.~\footnote{A sensitivity analysis of the impact of these parameters is discussed in supplementary materials.}
% A sensitivity analysis of the impact of these parameters is discussed later in Figures~\ref{Fig:pf_size}, \ref{Fig:pf_shape}, and~\ref{Fig:pf_weight}. 
% \arseni{in Figures or in the next section?}
% Justifications for the default values are presented later.
All algorithms were implemented in Python and ran on a MacBook Air with an Apple M1 chip and 8GB of RAM.

% \textbf{APFs with PIBT, LaCAM, and LaCAM$^*$}
%\paragraph{Negative Results in Standard MAPF}  In all our standard MAPF experiments, using APFs in PrP, LNS2, PIBT, and LaCAM, yielded either identical or inferior results.\footnote{A detailed description appears in the supplementary materials.} For PIBT, LaCAM, and LaCAM$^*$, we conjecture that this is due to the myopic nature of these algorithms, choosing a single step ahead in every iteration. For PrP and LNS2, we explain these poor results by the fact that the APFs encourage the single-agent path planning algorithm (TA* or SIPPS) to avoid the plans of agents that have already chosen a plan. Thus, the APFs only make the planning harder while avoiding plans that are already chosen to be part of the solution. Thus, the expected benefit of APFs ---  avoiding congested areas --- did not manifest in performance gains. 

\paragraph{Negative Results in Standard MAPF}  In all our standard MAPF experiments, using APFs in PrP, LNS2, PIBT, and LaCAM, yielded either identical or inferior results. We believe that the reason is as follows. 
In standard one-shot MAPF, APFs are less important because all we need is a single path for each agent that avoids other agents. 
Thus, avoiding conflicts (strong constraints) is enough to find a valid solution. 
Furthermore, for PrP and LNS2, when agent $a_i$ plans, then APFs only exist for agents that have already planned before $a_i$. 
% Thus, the effect of these APFs is marginal and did not show a significant advantage when used. 
Thus, the expected benefit of APFs ---  avoiding congested areas --- did not manifest in performance gains. 
However, finding plans that avoid congested areas can bring significant gains in the lifelong setting, where agents continuously receive new tasks over time. We demonstrate this in the next set of experiments, which evaluate our APF-augmented algorithms within the RCHR framework in a lifelong MAPF setting.

\paragraph{Lifelong MAPF}
% Next, we present the results of using APFs LMAPF setup~\cite{ma2017feasibility}.
% , where agents are assigned new goals on the fly. 
% We solved it under the RHCR framework where both $k=5$ and $w=5$, that is every 5 steps all agents need to replan and conflicts are only considered for the next 5 steps. 
% \arseni{repetition of the description of RHCR from the beginning of section 5.}
In this set of experiments, we solved LMAPF problems using the RHCR framework with the parameters \emph{window} and planning \emph{horizon} set to 5.
Each algorithm was limited by 10 seconds for the planning phase.
In planning-failure events, i.e., when an agent could not find a path within the time limit, we followed Morag et al.'s~\cite{fail_policies_2023} robust MAPF framework using the \emph{AllAgents} + \emph{iStay} + \emph{Persist} configuration. 
This corresponds to having all agents plan in every planning period (\emph{AllAgents}); 
failing agents stay in their place (\emph{iStay}); 
and all agents that do have a path and can follow it without conflicts, do so (\emph{Persist}). 
This configuration is simple to implement, commonly used in prior work, and yields reasonable results compared to other configurations according to Morag et al.~\cite{fail_policies_2023}. % (2023), who studied this issue. \roni{Bad sentence. } 
% In the planning phase, we executed all algorithms and our experiment halted after each agent performed 100 steps.
In the planning phase, our experiment halted after each agent performed 100 steps.
The main metric in the LMAPF experiments is the average \emph{throughput} of each algorithm, i.e., 
the number of times that an agent reaches its goals before the aforementioned 100 time-steps limit is reached, averaged over all agents.
Throughput is the common metric for measuring the quality of LMAPF algorithms \cite{RHCR_Li_Koenig_2021,song2023anytime,fail_policies_2023}.
%\footnote{The sum of costs and makespan are less relevant in the context of online MAPF where the flow of tasks is considered to be infinite.}
% Nevertheless, we do provide some results comparing the sum of costs (see Fig. \ref{fig:rsoc}).}

\begin{figure*} [!ht]
 \begin{center}
    \includegraphics[scale=0.25]{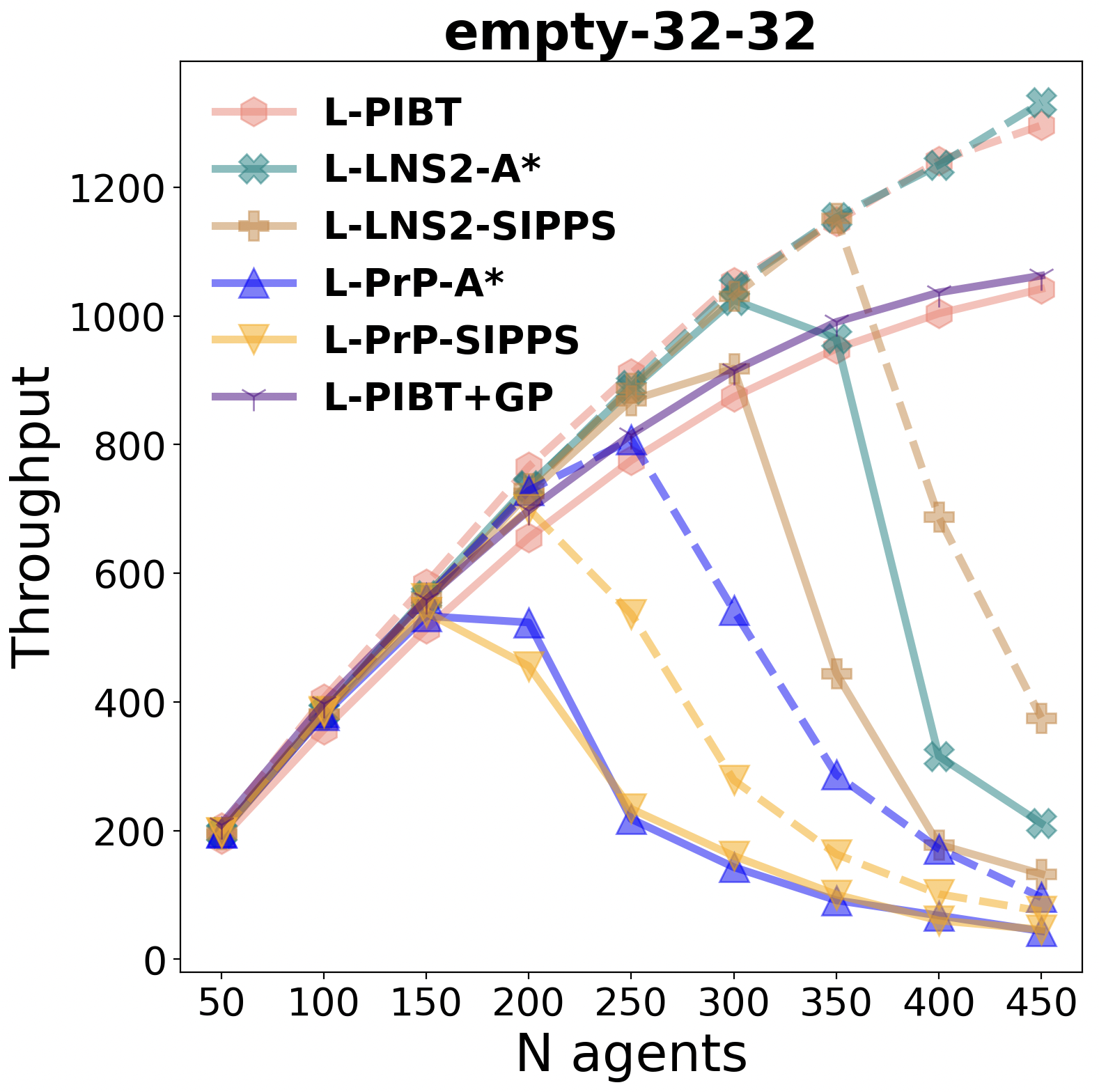}
  \includegraphics[scale=0.25]{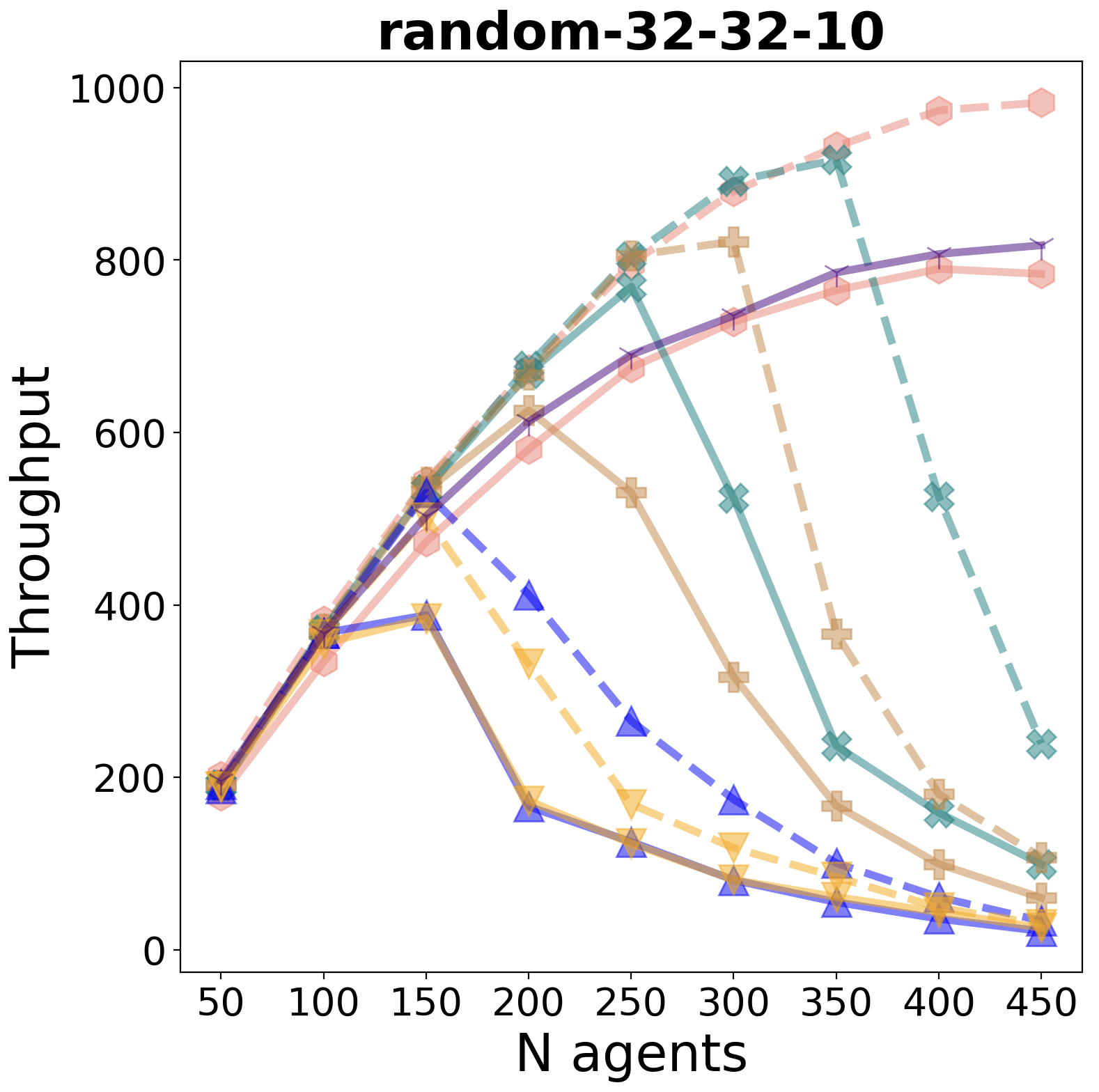}
  \includegraphics[scale=0.25]{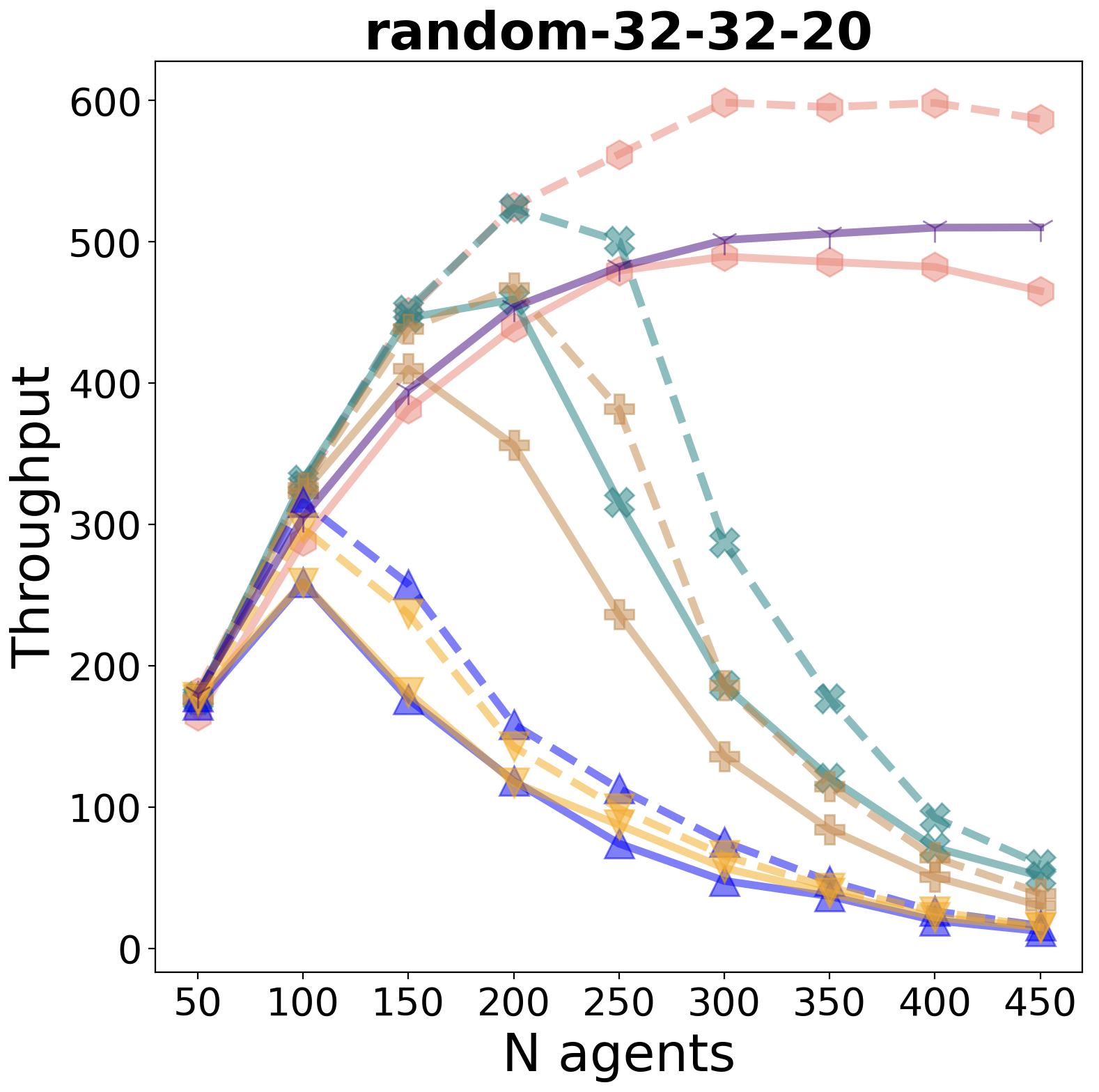}
  \includegraphics[scale=0.25]{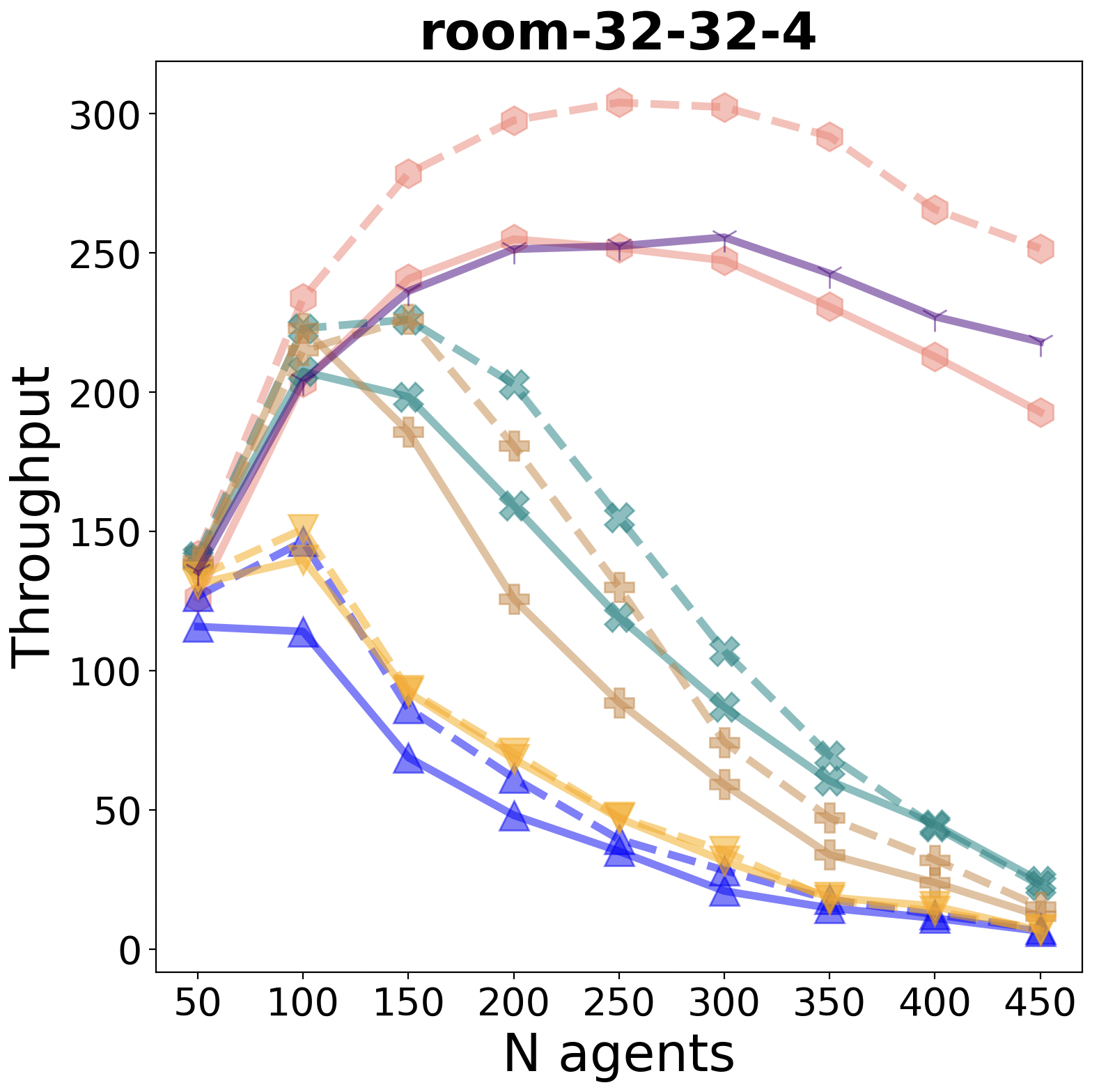}
 \caption{LMAPF: Average Throughput. Dashed lines - APF-enhanced; Solid lines - no APFs}
 \label{fig:throughput}
 % \vspace{-18pt}
 \end{center}
 \end{figure*}

Figure~\ref{fig:throughput} plots the average throughput of the different algorithms as a function of the number of agents. 
Solid lines represent original implementations of algorithms, and dashed lines represent our APF-enhanced versions of the same algorithms.
The results for LaCAM and LACAM$^*$ with and without APFs were virtually the same as for PIBT due to the fact that both LaCAM and LACAM$^*$ use PIBT as their low-level search algorithm. 
Therefore, we only plot the results of PIBT. 
The results clearly show that using APFs significantly increases the throughput of all other algorithms on all maps. 
For example, in the  \emph{empty-32-32} grid, LNS2 with \tapf\ reaches a throughput of around 1400 with 450 agents, which is approximately 7 times more than the throughput of vanilla LNS2 for the same number of agents. 
The significant advantage observed in using \tapf\ within the RHCR framework may suggest that it indeed achieves its intended purpose: biasing agents towards paths that avoid the paths of other agents, reducing future collisions and congestion. 
In most cases, PIBT+APF outperforms other approaches, including L-PIBT+GP, and improves the original PIBT by a significant margin. 
For example, in the \emph{room-32-32-4} grid, the throughput of PIBT+APF is higher by 20\% on average compared to PIBT.
% of  resolves collisions within the planning window, and by doing it with APF's assistance, RHCR generates the paths, where agents are far away from each other after the planning window.
% Thus, it helps to reduce future collisions and congestion.
As an interesting side phenomenon, we observed the superior performance of LNS2 algorithms using \astar, compared to LNS2 with SIPPS.
This could be due to the fact that \astar\ prioritizes short paths and SIPPS prioritizes paths with minimum soft constraints.
% Our APF-enhanced version of PIBT also significantly outperformed L-PIBT+GP across all the grids in our experiments.

To summarize, while APFs did not improve performance for standard MAPF they were found to be very helpful in LMAPF for all algorithms. In LMAPF agents need to continuously plan new paths. Thus, avoiding congested areas is very important and APFs play a significant role in achieving this.

%%%%%%%%%%%%%%%%%%%%%%%%%%%%%%%%%%%%%%
%%%%%%%%%%%%%%%%%%%%%%%%%%%%%%%%%%%%%%
%%%%%%%%%%%%%%%%%%%%%%%%%%%%%%%%%%%%%%
%%%%%%%%%%%%%%%%%%%%%%%%%%%%%%%%%%%%%%
%%%%%%%%%%%%%%%%%%%%%%%%%%%%%%%%%%%%%%
%%%%%%%%%%%%%%%%%%%%%%%%%%%%%%%%%%%%%%
%%%%%%%%%%%%%%%%%%%%%%%%%%%%%%%%%%%%%%
%%%%%%%%%%%%%%%%%%%%%%%%%%%%%%%%%%%%%%
%%%%%%%%%%%%%%%%%%%%%%%%%%%%%%%%%%%%%%
%%%%%%%%%%%%%%%%%%%%%%%%%%%%%%%%%%%%%%
%%%%%%%%%%%%%%%%%%%%%%%%%%%%%%%%%%%%%%
\paragraph{Parameter Sensitivity Analysis}
% Parameters:
% size - how far away from the location do the potential field reach  - 4
% shape - how the potential fields spreads within the "size" - 2
% weight - how much a potential can add to a step - 1

Our APF algorithms require setting several parameters, \maxdist, \shape, $w$, and $t_{max}$. 
Next, we present a sensitivity analysis for these parameters.
As a representative example, we focus on the algorithm that performs the best in our experiments, PIBT+APF.
We show the impact of the PIBT+APF parameters, \maxdist, \shape, $w$, and $t_{max}$ on the overall performance in our LMAPF experiments. 
We only report results on the \emph{room-32-32-4} map, but similar trends were observed with other algorithms and other maps. 

 \begin{figure*} [!ht]
 \begin{center}
  % \subfloat[$w$ values]{\includegraphics[scale=0.158]{pics/params_pibt_w_1.png}}
  % \subfloat[$\maxdist$ values]{\includegraphics[scale=0.158]{pics/params_pibt_d_max_1.png}}
  % \subfloat[$\shape$ values]{\includegraphics[scale=0.158]{pics/params_pibt_gamma_1.png}}
  % \subfloat[$t_{max}$ values]{\includegraphics[scale=0.158]{pics/params_pibt_t_max_1.png}}
    \subfloat[$w$ values]{\includegraphics[scale=0.24]{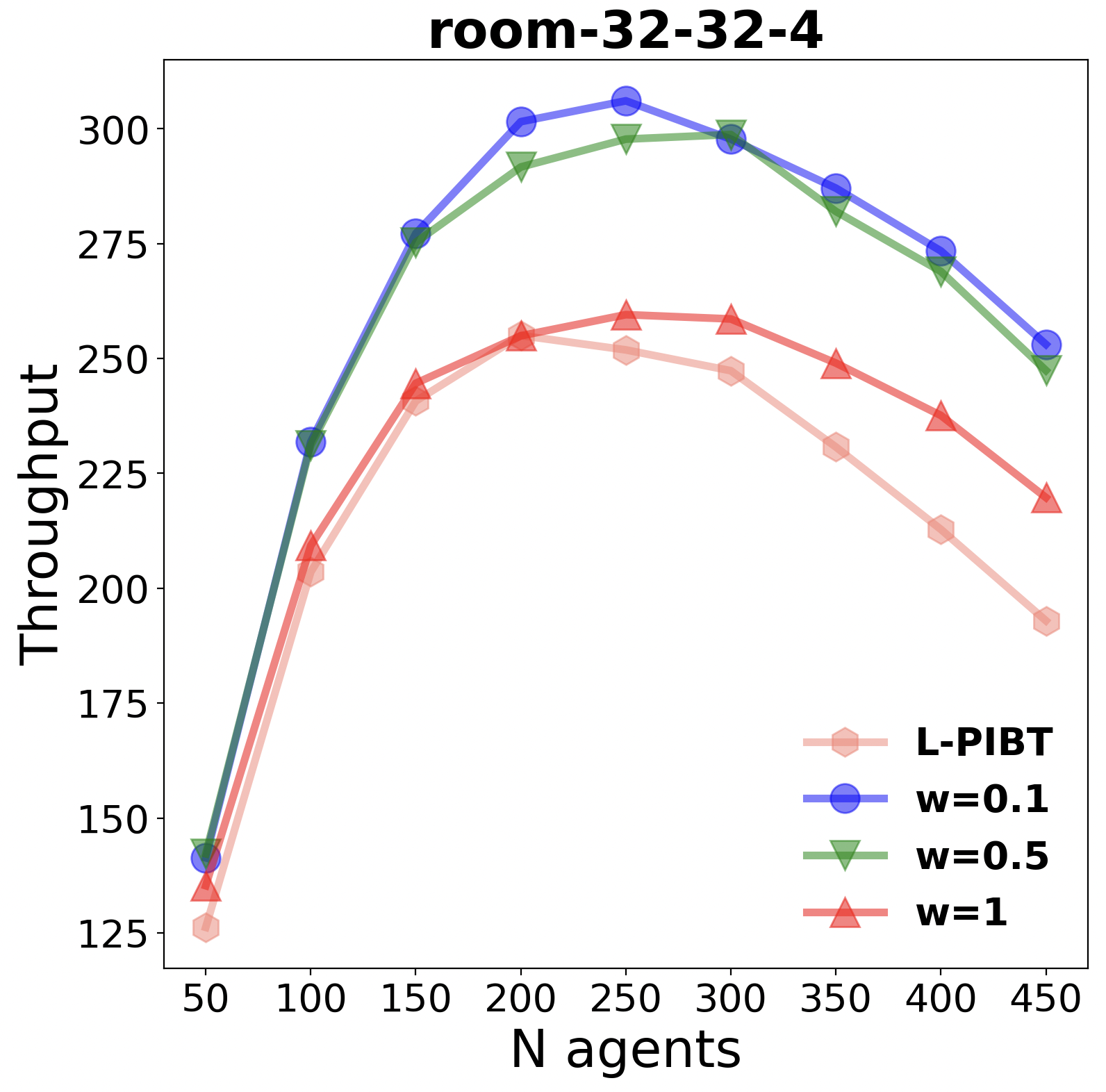}}
  \subfloat[$\maxdist$ values]{\includegraphics[scale=0.24]{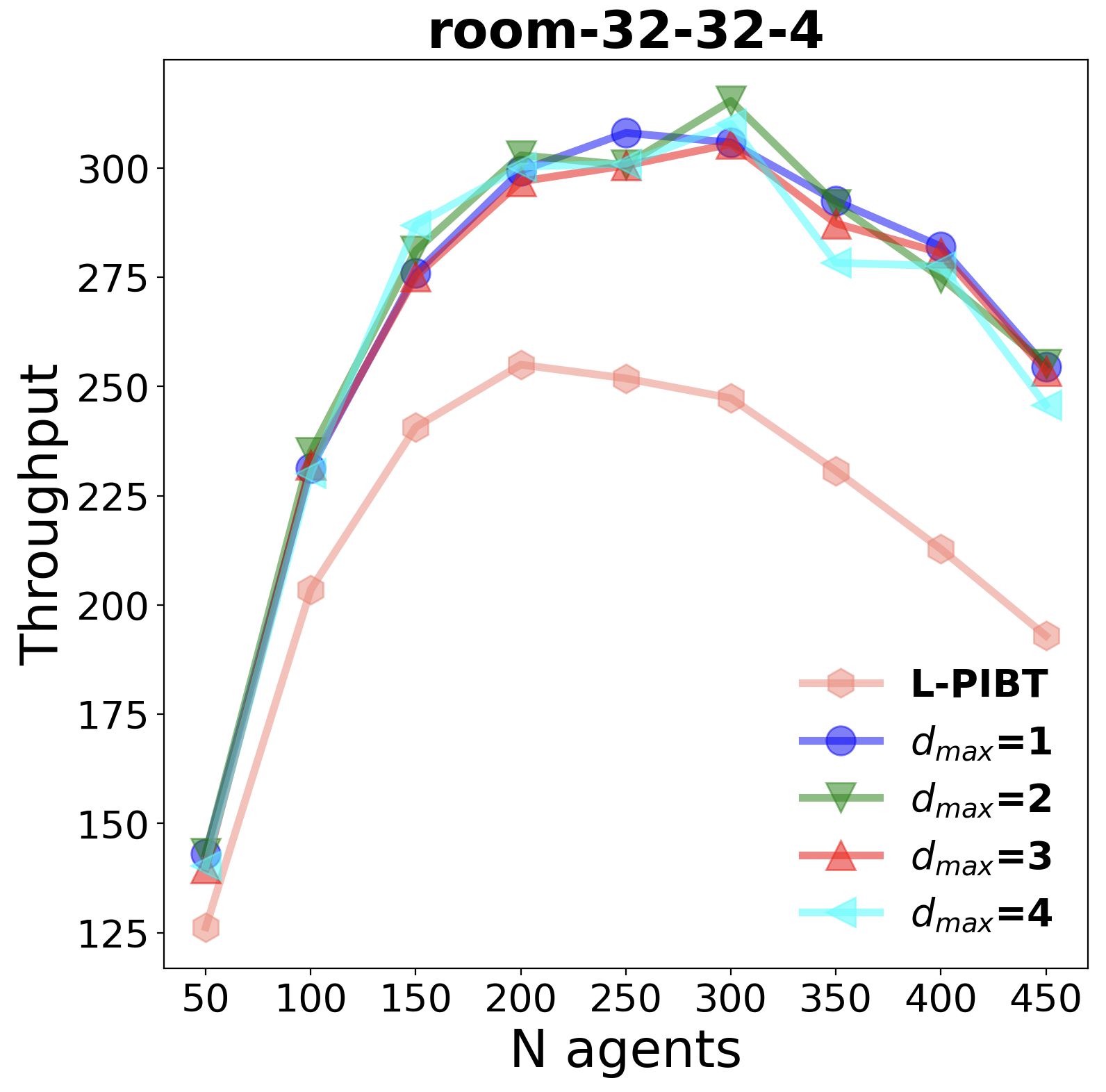}}
  
  \subfloat[$\shape$ values]{\includegraphics[scale=0.24]{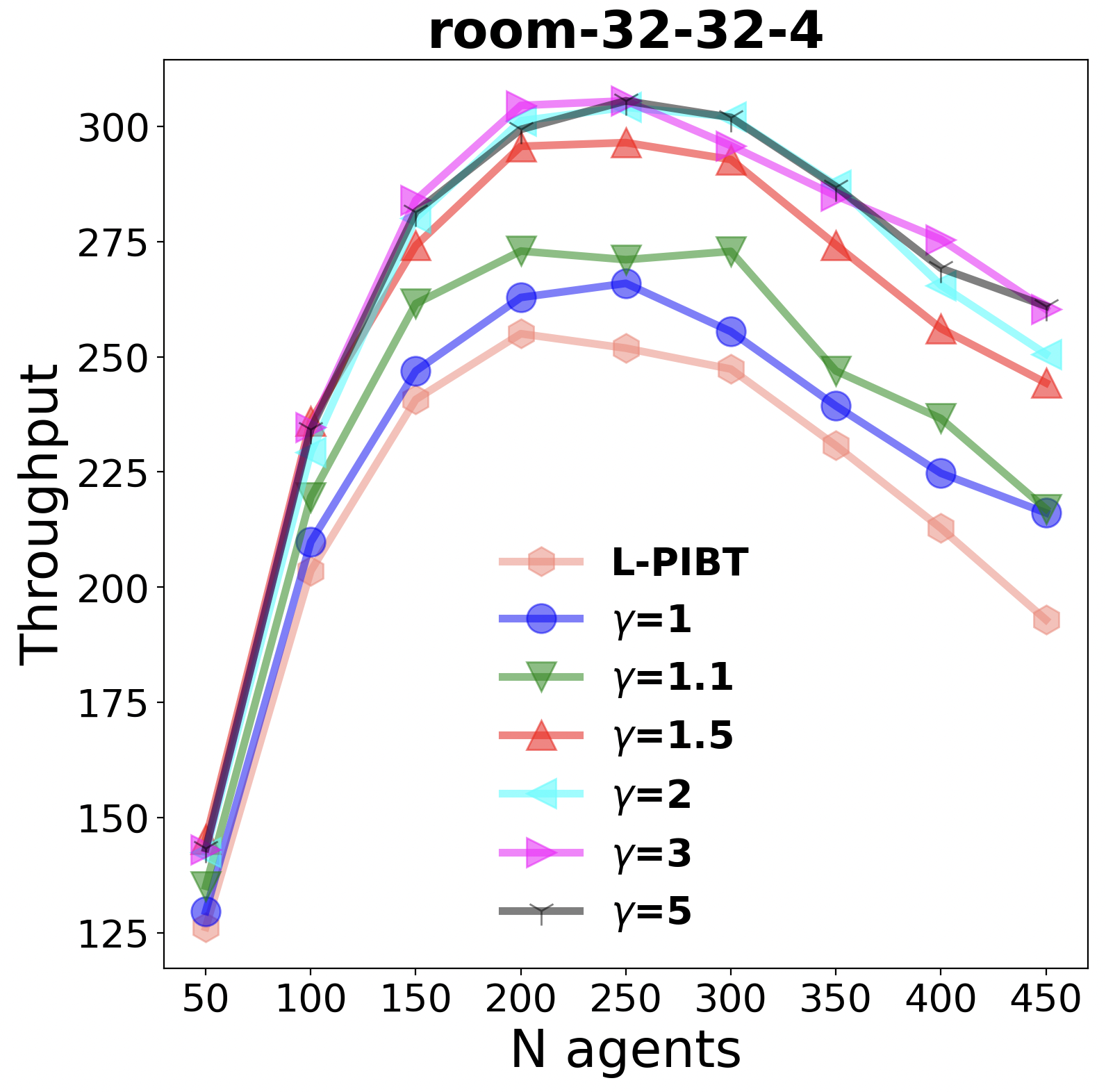}}
  \subfloat[$t_{max}$ values]{\includegraphics[scale=0.24]{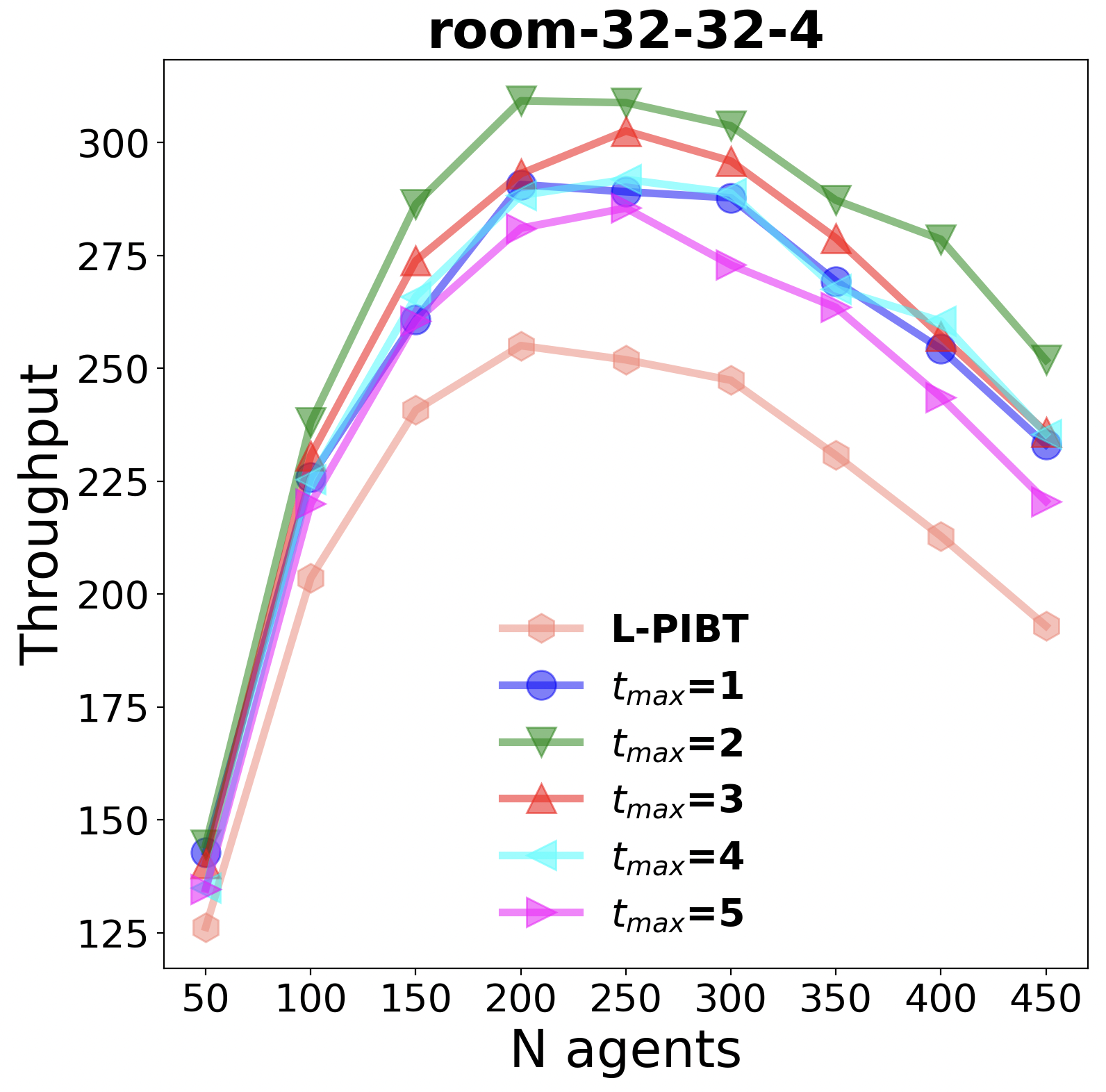}}
 \caption{Parameter sensitivity analysis.
 The ``L-PIBT" label designates the original PIBT algorithm that does not utilize APFs.}
 \label{Fig:params_apf}
 \end{center}
 \end{figure*}
 
Figure~\ref{Fig:params_apf} plots the average throughput as a function of the number of agents for different values of the parameters. 
Consider the impact of the $w$ parameter (Fig.~\ref{Fig:params_apf}(a)), which determines how much weight should be given to the cost resulting from APFs as opposed to the regular cost of moving. 
The impact of the value of this parameter seems to be crucial as different values provide very different results. 
For example, $w=0.1$ yielded the best results, while setting $w=1$ was only slightly better than PIBT (where $w=0$).

Next, consider the impact of varying $\maxdist$ (Fig.~\ref{Fig:params_apf}(b)). Recall that $\maxdist$ defines how far away from an agent's planned location a potential field reaches. 
The results show that setting $\maxdist$ to either extreme value, too small (\maxdist=1) or too high (\maxdist=4), yields very similar results. We chose \maxdist=2, as it was the one to yield the best performance in all other grids.

Fig.~\ref{Fig:params_apf}(c) analyzes the impact of \shape , which defines how fast the impact of the APF decreases with the distance from its origin. Setting $\shape=1$ corresponds to an APF that has a uniform effect regardless of distance, as long as the distance is smaller than \maxdist. 
This assignment of $\shape=1$ yielded similar results to vanilla PIBT. All other values of \shape\ yielded much better results, where $\shape=2$ worked best in all settings. 

Finally, consider the $t_{max}$ (Fig.~\ref{Fig:params_apf}(d)), which defines how many steps forward PIBT+APF builds a future path that ignores other agents and creates APF around the path.
Interestingly, $t_{max}=2$ yielded significantly better results than all other values, neither lower nor greater than 2.

\section{Conclusion and Future Work}

We investigated whether MAPF can be solved efficiently using artificial potential fields (APFs). 
First, we showed that a direct implementation of APFs in a myopic manner is fast but may yield poor results. 
Then, we proposed to use APFs in \ta\ and in SIPPS, which are key components of many MAPF algorithms. 
Specifically, we incorporated APFs in the calculation of the $g$ component of \ta's node evaluation function and the $c$ and $g$ components of SIPPS's node evaluation functions.
Next, we proposed a way to incorporate APFs into the PIBT and LaCAM algorithms.
The resulting algorithms add bias to the search to avoid passing near the paths of other agents.

Experimentally, we showed that the introduction of APFs in other algorithms did not yield any advantage when solving a single offline MAPF problem. 
However, in the context of lifelong MAPF, we showed that using APFs in \ta, SIPPS, and PIBT is highly beneficial, resulting in significantly higher overall system throughput. 

There are several interesting directions for future work.
One such direction is to study how to effectively set the parameters \maxdist, \shape, $t_{max}$, and $w$ for APF in \ta, SIPPS and PIBT. 
Another direction is to examine better ways to incorporate APFs to solve one-shot MAPF problems.
%%%%%%%%%%%%%%%%%%%%%%%%%%%%%%%%%%%%%%%%%%%%%%%%%%%%%%%%%%%%%%%%%%%%%%%%
%%%%%%%%%%%%%%%%%%%%%%%%%%%%%%%%%%%%%%%%%%%%%%%%%%%%%%%%%%%%%%%%%%%%%%%%
%%%%%%%%%%%%%%%%%%%%%%%%%%%%%%%%%%%%%%%%%%%%%%%%%%%%%%%%%%%%%%%%%%%%%%%%
%%%%%%%%%%%%%%%%%%%%%%%%%%%%%%%%%%%%%%%%%%%%%%%%%%%%%%%%%%%%%%%%%%%%%%%%
%%%%%%%%%%%%%%%%%%%%%%%%%%%%%%%%%%%%%%%%%%%%%%%%%%%%%%%%%%%%%%%%%%%%%%%%
%%%%%%%%%%%%%%%%%%%%%%%%%%%%%%%%%%%%%%%%%%%%%%%%%%%%%%%%%%%%%%%%%%%%%%%%
\bibliography{sample}

%%%%%%%%%%%%%%%%%%%%%%%%%%%%%%%%%%%%%%%%%%%%%%%%%%%%%%%%%%%%%%%%%%%%%%%%
%%%%%%%%%%%%%%%%%%%%%%%%%%%%%%%%%%%%%%%%%%%%%%%%%%%%%%%%%%%%%%%%%%%%%%%%
%%%%%%%%%%%%%%%%%%%%%%%%%%%%%%%%%%%%%%%%%%%%%%%%%%%%%%%%%%%%%%%%%%%%%%%%
\appendix

\section{Justification for Parameters}

\begin{figure*} [t]
%\vspace{-8pt}
  % \hspace{-0pt}
 \begin{center}
  \subfloat[Different $w$ values]{\includegraphics[scale=0.25]{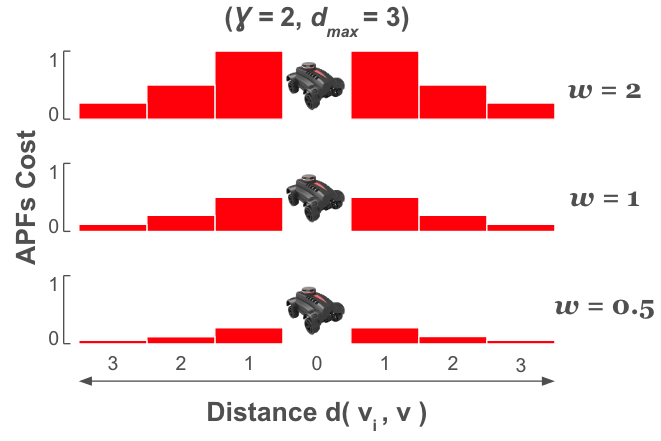}}
  \subfloat[Different $\gamma$ values]{\includegraphics[scale=0.25]{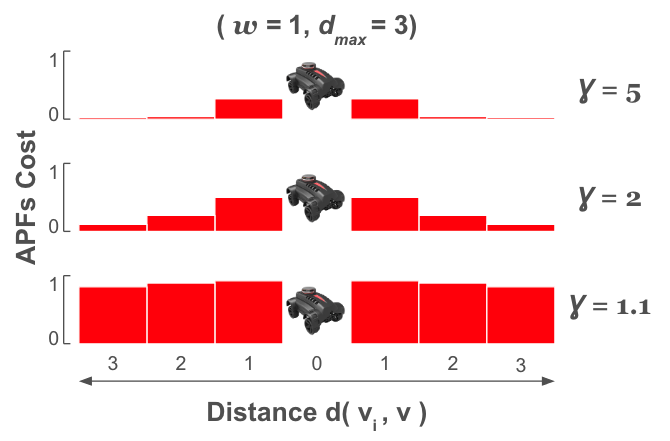}}
  \subfloat[Different $d_{max}$ values]{\includegraphics[scale=0.25]{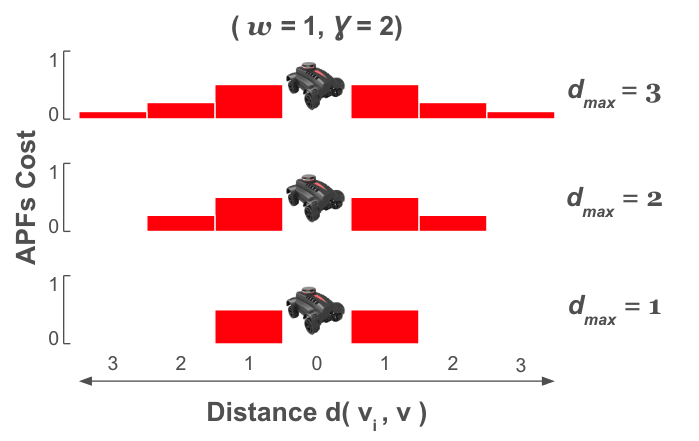}}
% \vspace{-10pt}
 \caption{The influence of parameters' values on APFs. (a) $w$ controls the strength; (b) $\gamma$ controls the rate of decay; (c) $d_{max}$ defines the radius of influence.}
 \label{fig:apfs_params}
 % \vspace{-18pt}
 \end{center}
 \end{figure*}
To demonstrate the role of each parameter, consider Figure \ref{fig:apfs_params}. 
The $x$-axis is the distance from the agent (in the middle), the $y$-axis is the value of APFs, and the height of the red bars represents the specific APFs cost for the locations of the specific distance from the agent.
All parameters of APFs are constant, except those we want to stir a bit to see the impact.
We stir each parameter separately.
We can clearly see that the $w$ parameter controls the strength of the repulsion APF (Fig. \ref{fig:apfs_params}(a)), $\gamma$ controls the rate of decay, i.e. how fast its intensity declines while moving away from its source (Fig. \ref{fig:apfs_params}(b)), and $d_{max}$ defines how far away from $v_i$ the repulsion APF will influence the cost (Fig. \ref{fig:apfs_params}(c)).

\section{Supplementary Sensitivity Analysis}

In addition to the examination of \tapf\ parameters in the main paper, we present a similar analysis of \sippsapf and \pibtapf parameters.
Recall, that as a representative example, we focus on the algorithm with the best performance in our experiments --- LNS2 with \tapf.
Here, we present the results only with \emph{random-32-32-10} grid, however, analogous trends were observed in other algorithms and other grids as well. 

\subsection{\sippsapf}

 \begin{figure*} [!ht]
%\vspace{-8pt}
   % \hspace{2pt}
 \begin{center}
  \subfloat[$\maxdist$ values]{\includegraphics[scale=0.17]{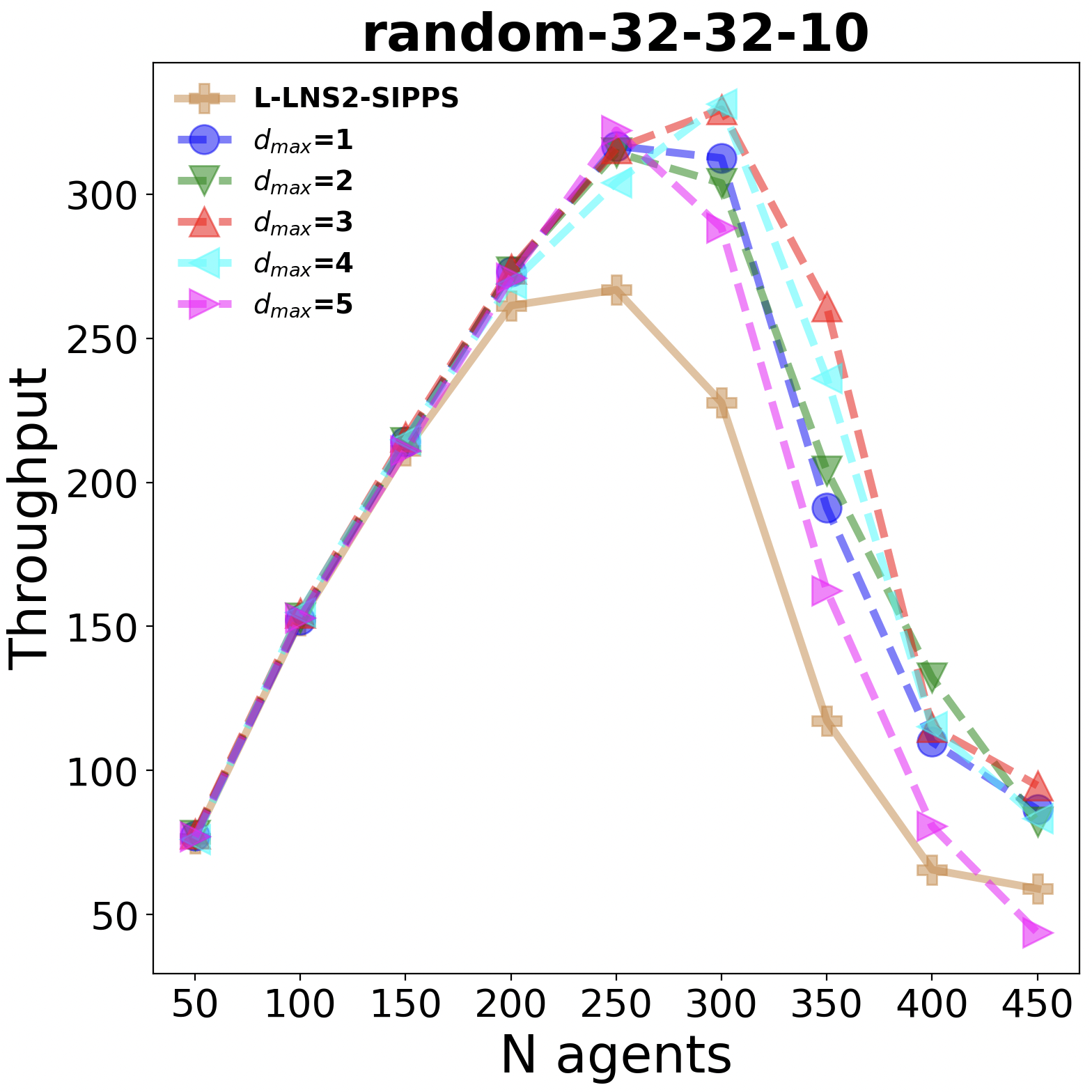}}
  \hspace{15pt}
  \subfloat[$\shape$ values]{\includegraphics[scale=0.17]{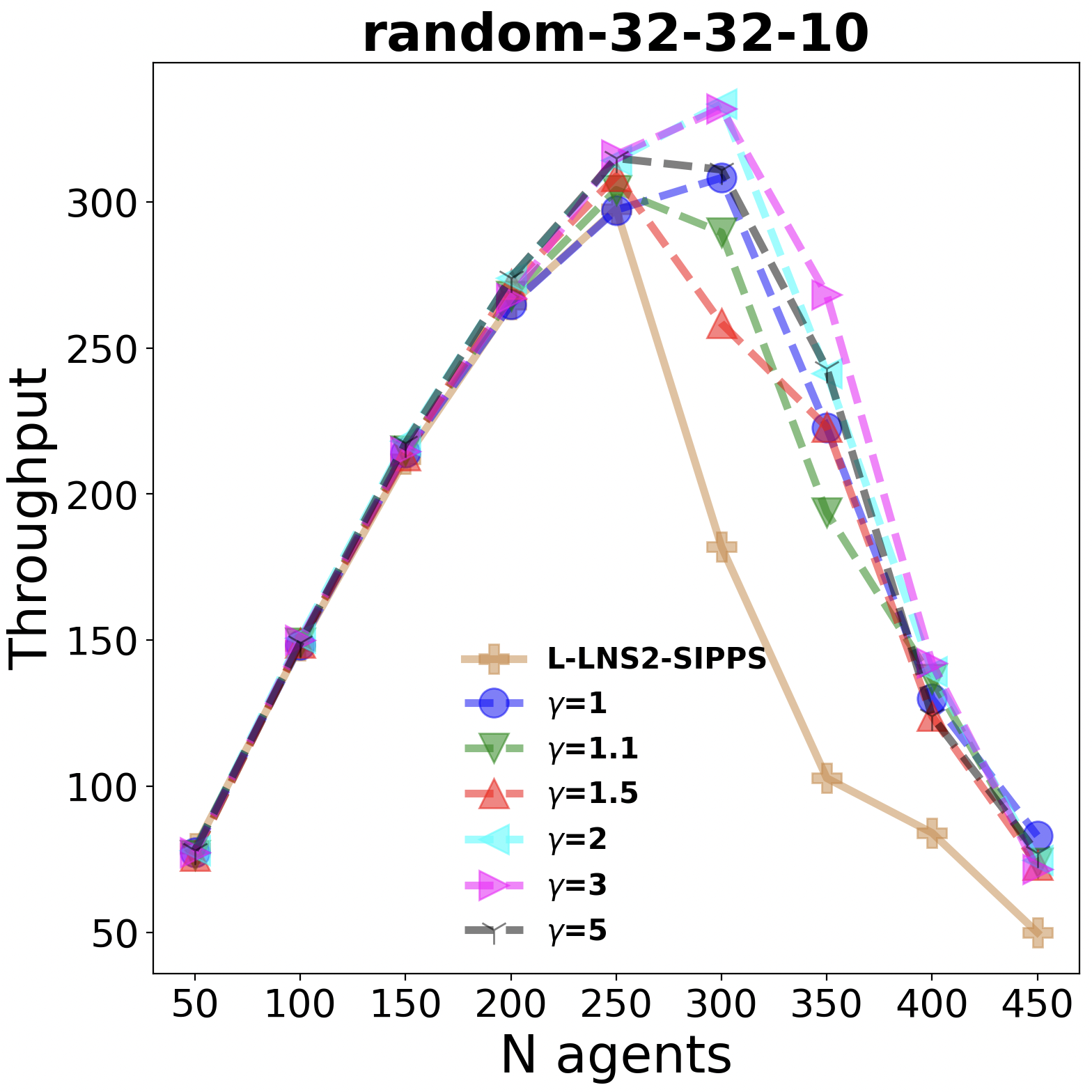}}
  \hspace{15pt}
  \subfloat[$w$ values]{\includegraphics[scale=0.17]{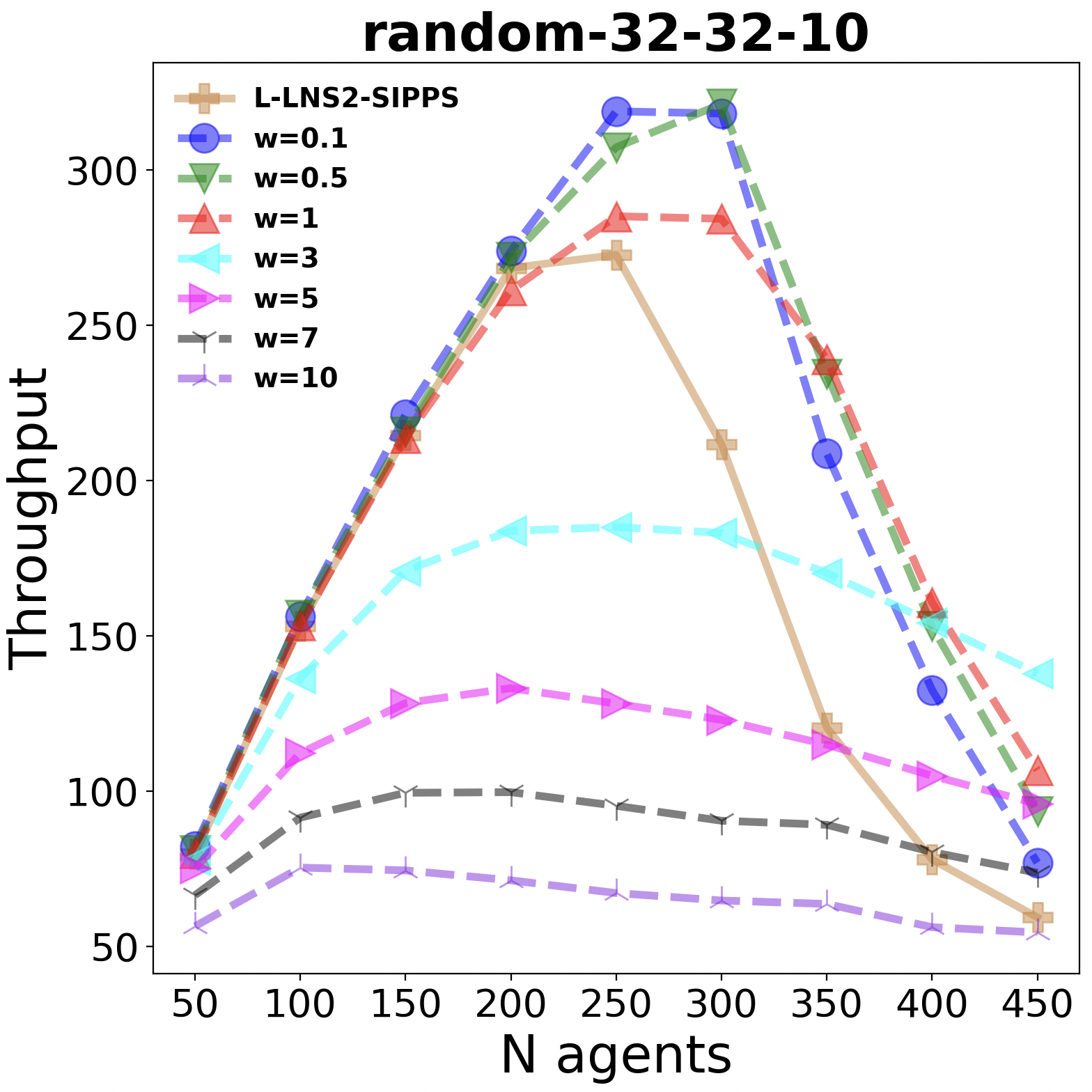}}
 \caption{Throughput for different parameter values of APFs for LNS2 with \tapf\ .}
 \label{Fig:params_sipps_apf}
 % \vspace{-18pt}
 \end{center}
 \end{figure*}

Figure~\ref{Fig:params_sipps_apf} plots an average throughput as a function of the number of agents, for different values of $\maxdist$ (Fig. \ref{Fig:params_sipps_apf}(a)), $\shape$ (Fig. \ref{Fig:params_sipps_apf}(b)), and $w$ (Fig. \ref{Fig:params_sipps_apf}(c)). 
Regarding $\maxdist$ and $\shape$, almost all the values resulted in better throughput, when values $\maxdist=3$ and $\shape=3$ were of the best performance.
The trend in $w$ was different. For low values, the results were almost always better than a vanilla version. For high values of $w$ the results were inferior to almost every number of agents, except the dense scenarios, where, for example, $w=3$ succeeded in reaching the highest throughput.
In our experiments, we chose $w=0.1$.

\subsection{\tapf}

 \begin{figure*} [!ht]
%\vspace{-8pt}
   % \hspace{2pt}
 \begin{center}
  \subfloat[$\maxdist$ values]{\includegraphics[scale=0.17]{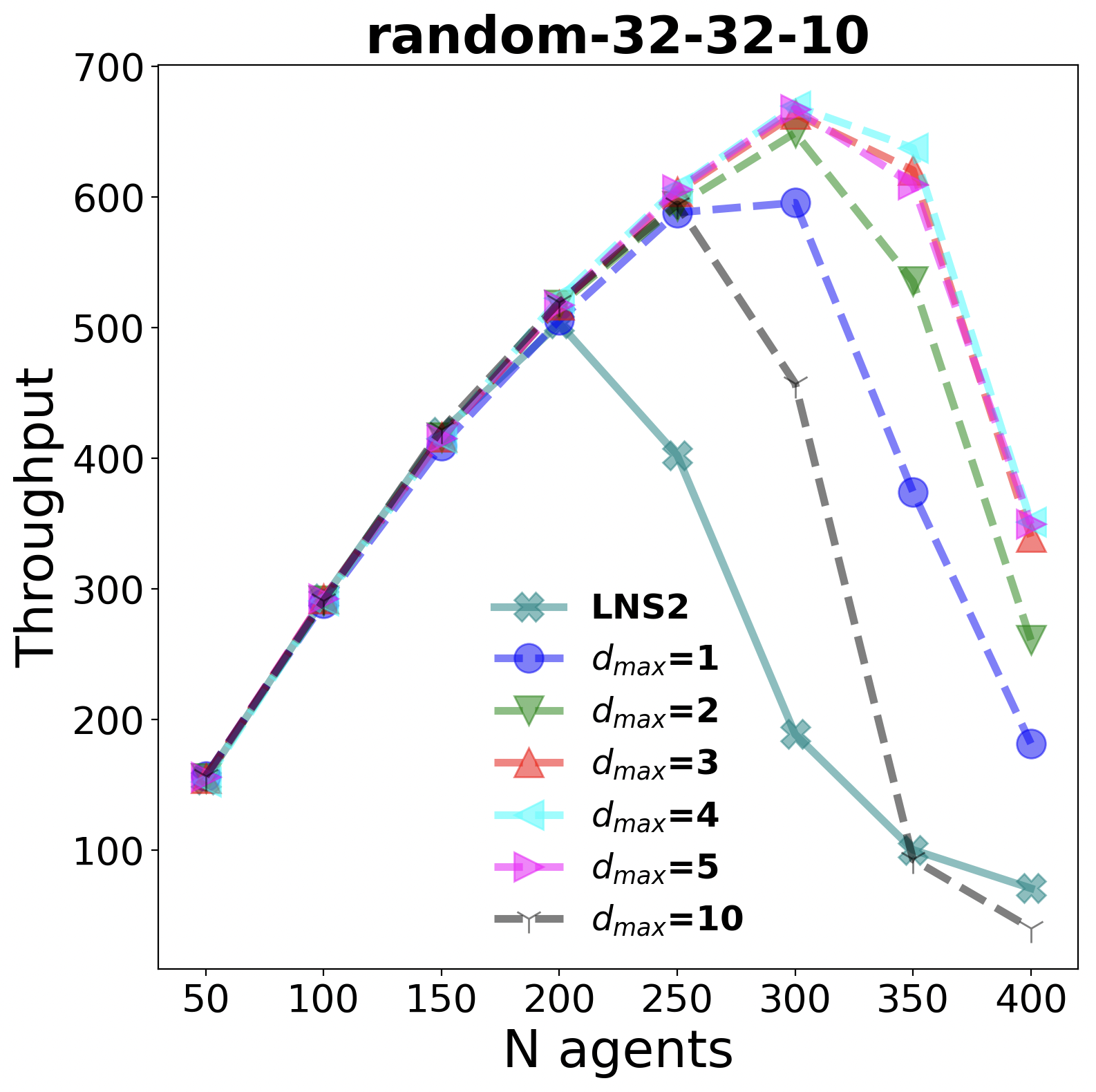}}
  \hspace{15pt}
  \subfloat[$\shape$ values]{\includegraphics[scale=0.17]{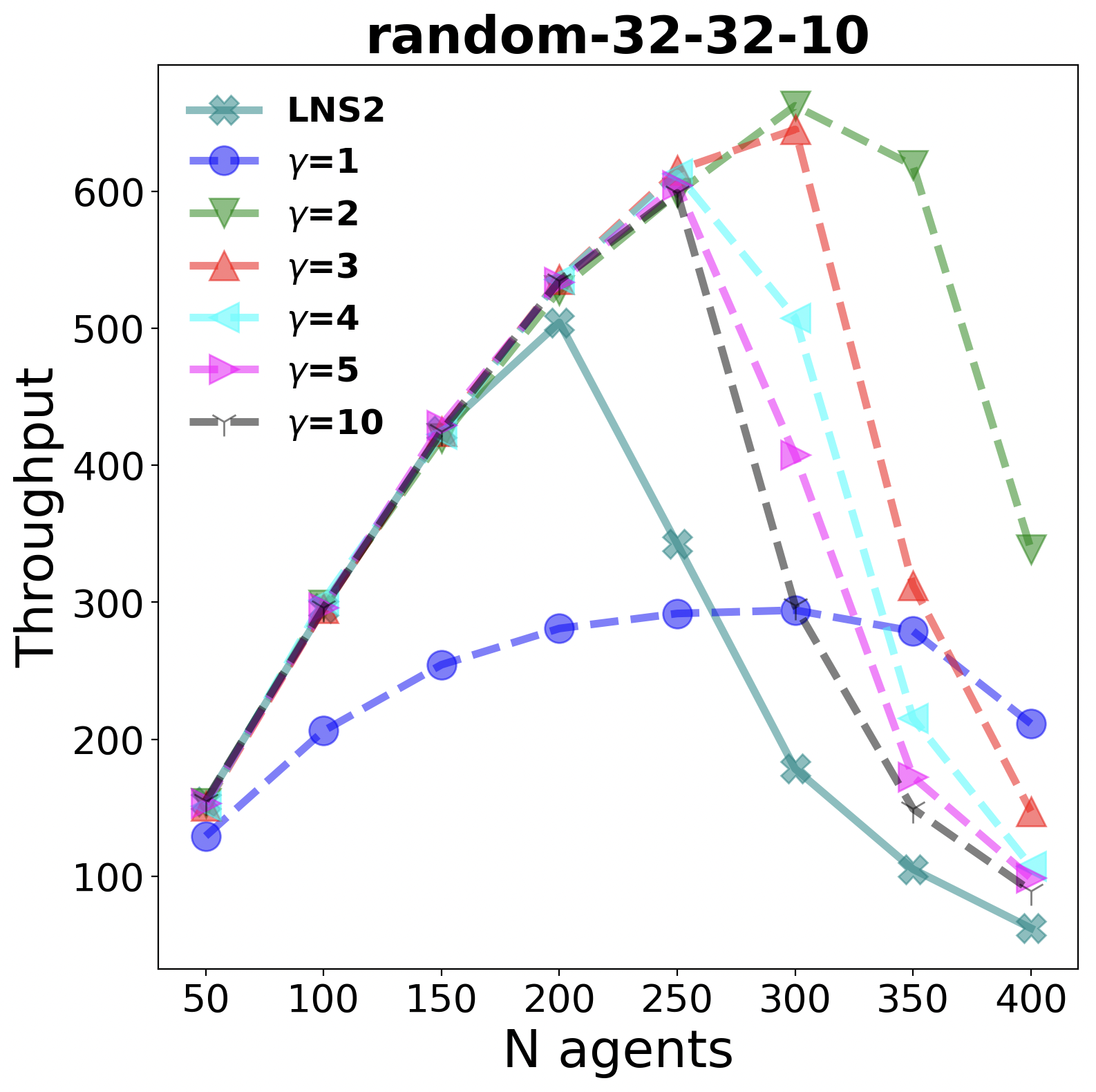}}
  \hspace{15pt}
  \subfloat[$w$ values]{\includegraphics[scale=0.17]{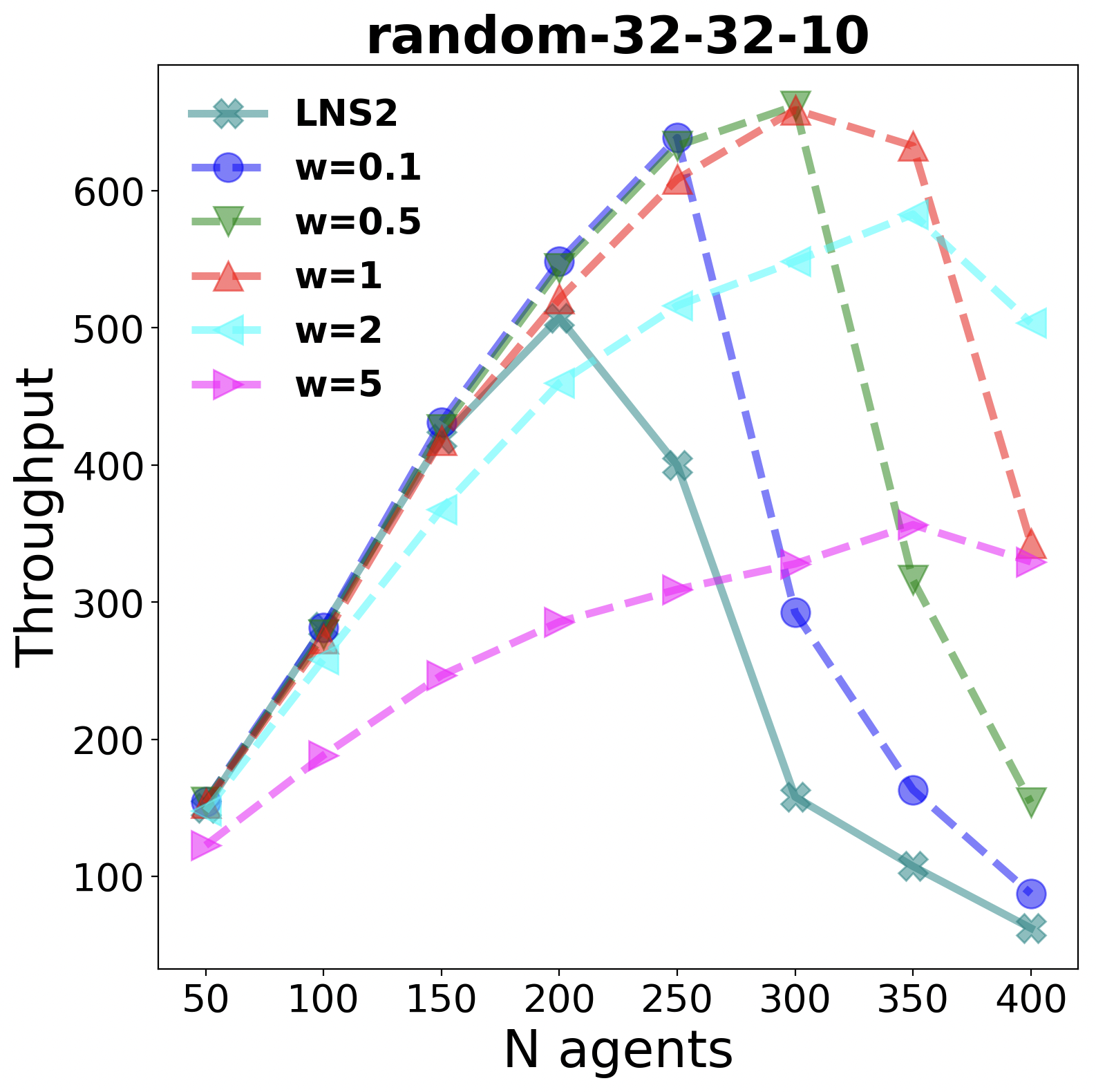}}
 \caption{Throughput for different parameter values of APFs for LNS2 with \tapf\ .}
 \label{Fig:params_apf}
 % \vspace{-18pt}
 \end{center}
 \end{figure*}

 We show the impact of the \tapf\ parameters, \maxdist, \shape, and $w$, on the overall performance in our LMAPF experiments. 
We only report results on the \emph{random-32-32-10} map but similar trends were observed with other algorithms and other maps. 

Figure~\ref{Fig:params_apf} plots the average throughput as a function of the number of agents, for different values of $\maxdist$, $\shape$, and $w$. 
Consider the impact of varying $\maxdist$ (Fig.~\ref{Fig:params_apf}(a)). Recall that $\maxdist$ defines how far away from an agent's planned location a potential field reaches. 
The results show that setting $\maxdist$ to either extreme value, too small (\maxdist=1) or too high (\maxdist=10), yields significantly worse results than setting $\maxdist$ to 4. Yet, even in these cases, APF-enhanced LNS2 significantly outperformed plain LNS2.

Next, consider  \shape\ (Fig.~\ref{Fig:params_apf}(b)), which defines how fast the impact of the APF decreases with the distance from its origin. Setting $\shape=1$ corresponds to an APF that has a uniform effect regardless of distance, as long as the distance is smaller than \maxdist. 
This assignment of $\shape=1$ yielded even worse results than vanilla LNS2. All other values of \shape\ yielded much better results, where \shape\ =2 worked best in this setting. 

Finally, consider the impact of the $w$ parameter (Fig.~\ref{Fig:params_apf}(c)), which determines how much weight should be given to the cost stemming from APFs as opposed to the regular cost of moving. 
The impact of the value of this parameter is interesting because its preferred value depends on the number of agents in the problem. 
For example, $w=2$ yielded the best results for problems with 400 agents, while setting $w=1$ was best for the other.
We also explored setting different $w$ values for different agents according to their traveled path length or future path lengths. These methods did not find any significant improvements, and thus their results are not reported.

\section{MAPF Experiments}
\begin{figure} [ht]
 \begin{center}
  \includegraphics[scale=0.5]{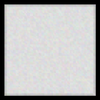}
  \includegraphics[scale=0.5]{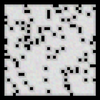}
  \includegraphics[scale=0.5]{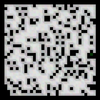}
  \includegraphics[scale=0.5]{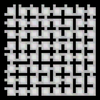}
% \vspace{-10pt}
 \caption{MAPF Grids}
 \label{fig:maps}
 \end{center}
 \end{figure}
In this section, we present results in a standard MAPF setting. 
The implementation of APFs without the RHCR~\cite{li2021anytime} framework (i.e. original algorithms) yields inferior results.
This can be described by the fact that the first agents do not consider others that come afterward and, therefore cannot take advantage of their APFs to, potentially, escape congestion.
Whereas in RHCR, the agents may reconsider several times their paths before arriving at a goal.
Hence, we omit the results without RHCR.
For the rest of the section, all PrP and LNS2 algorithms are implemented within RHCR, where the window size and horizon depth were set to 5, which we found to be effective in our experimental setup.
In this set of experiments, a time limit of one minute was imposed. 
For the sake of clarity, we do not report on APF-enhanced versions of PIBT, LaCAM, and LaCAM$^*$, as they showed no substantial difference from their baseline versions.
As we mentioned in the main paper, we conjecture that this is due to the myopic structure of these algorithms, choosing a single step ahead in every iteration. 
All experiments were performed on four different maps from the MAPF benchmark~\cite{stern2019mapf}: \emph{empty-32-32}, \emph{random-32-32-10}, \emph{random-32-32-20}, and \emph{room-32-32-4}, as they present different levels of difficulty. 
The maps are visualized in the plots in Figure~\ref{fig:maps}. 

\paragraph{Success Rate}
\begin{figure*} [!ht]
    \begin{center}
        \includegraphics[scale=0.15]{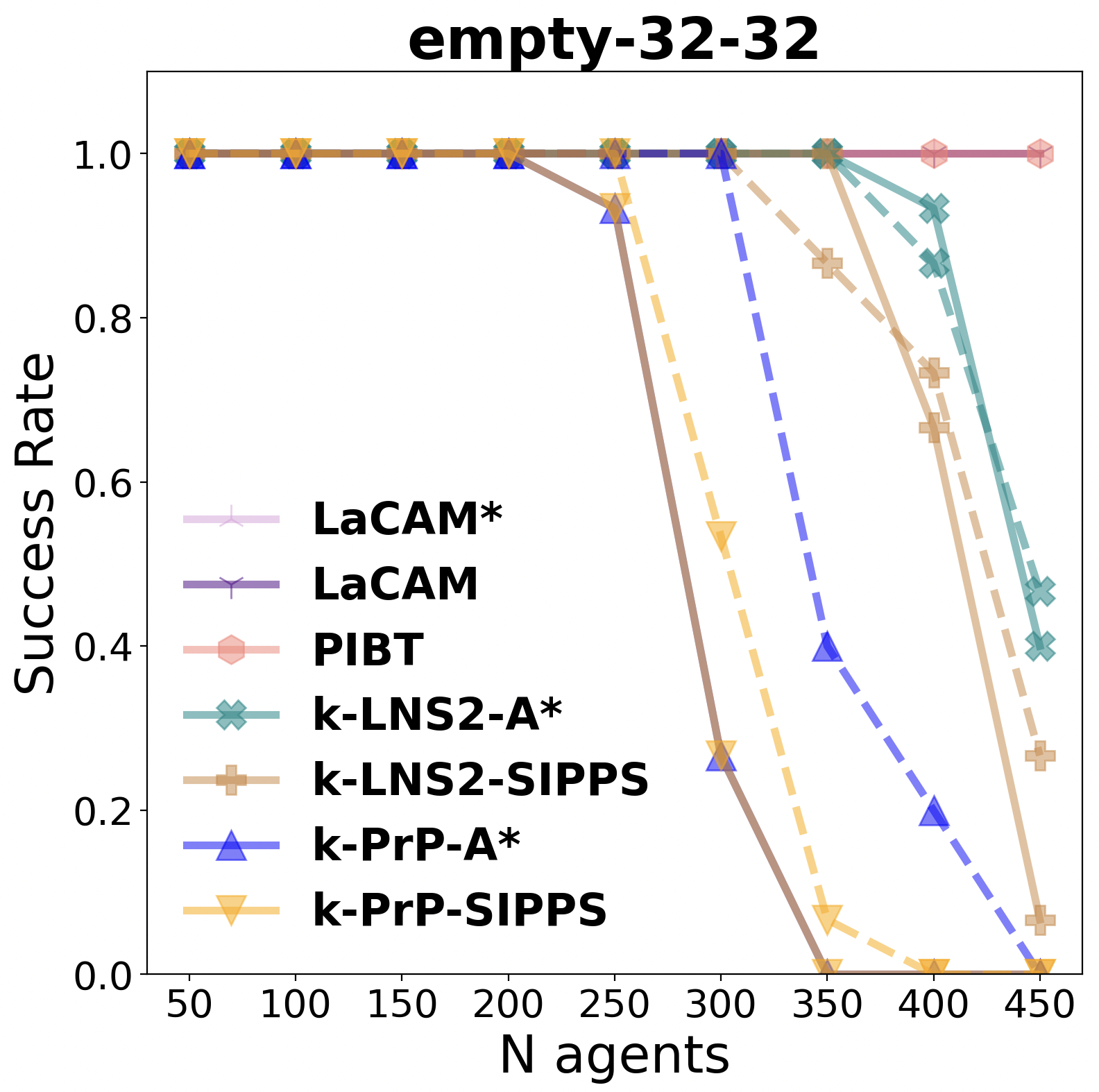}
        \includegraphics[scale=0.15]{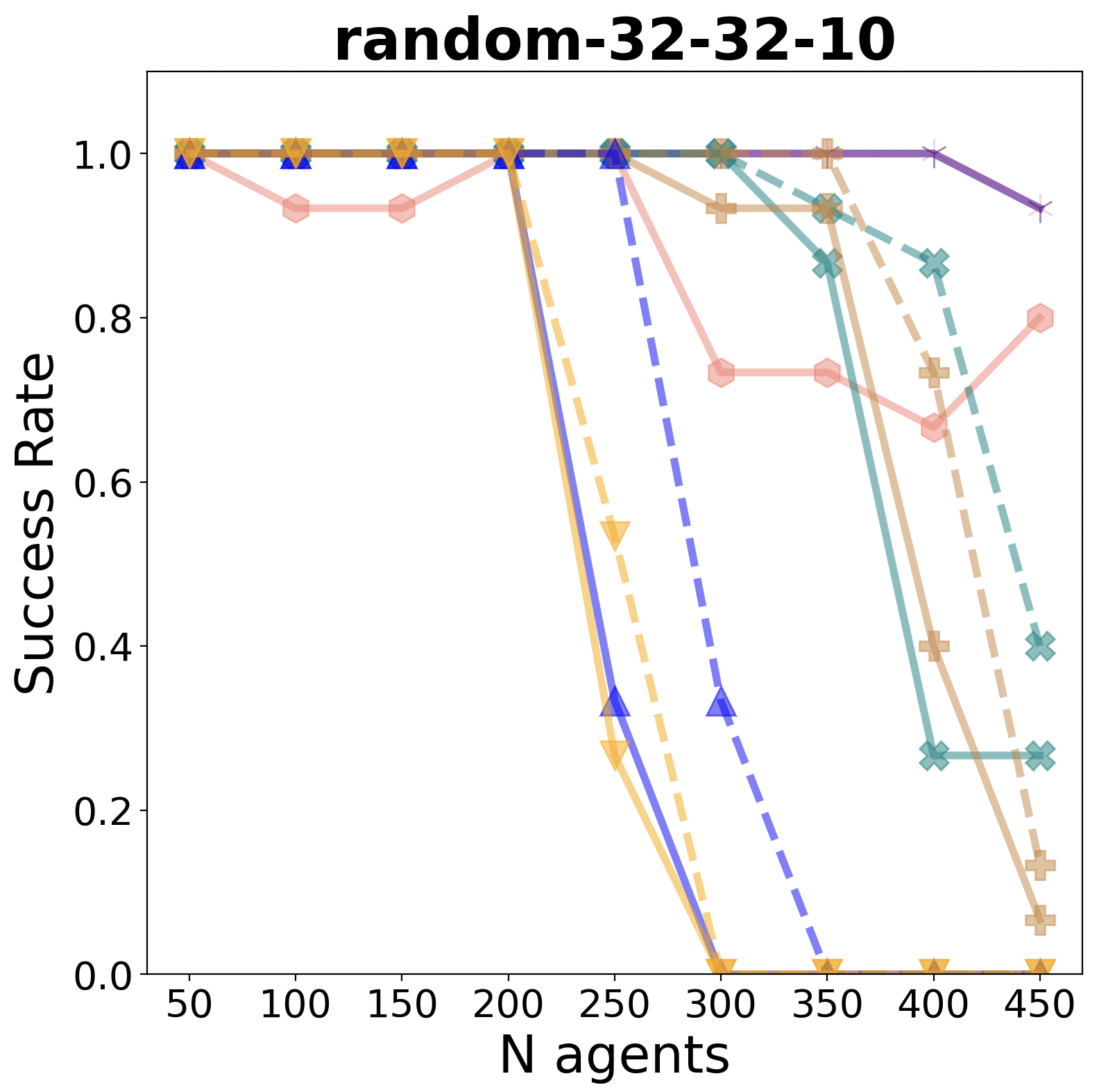}
        \includegraphics[scale=0.15]{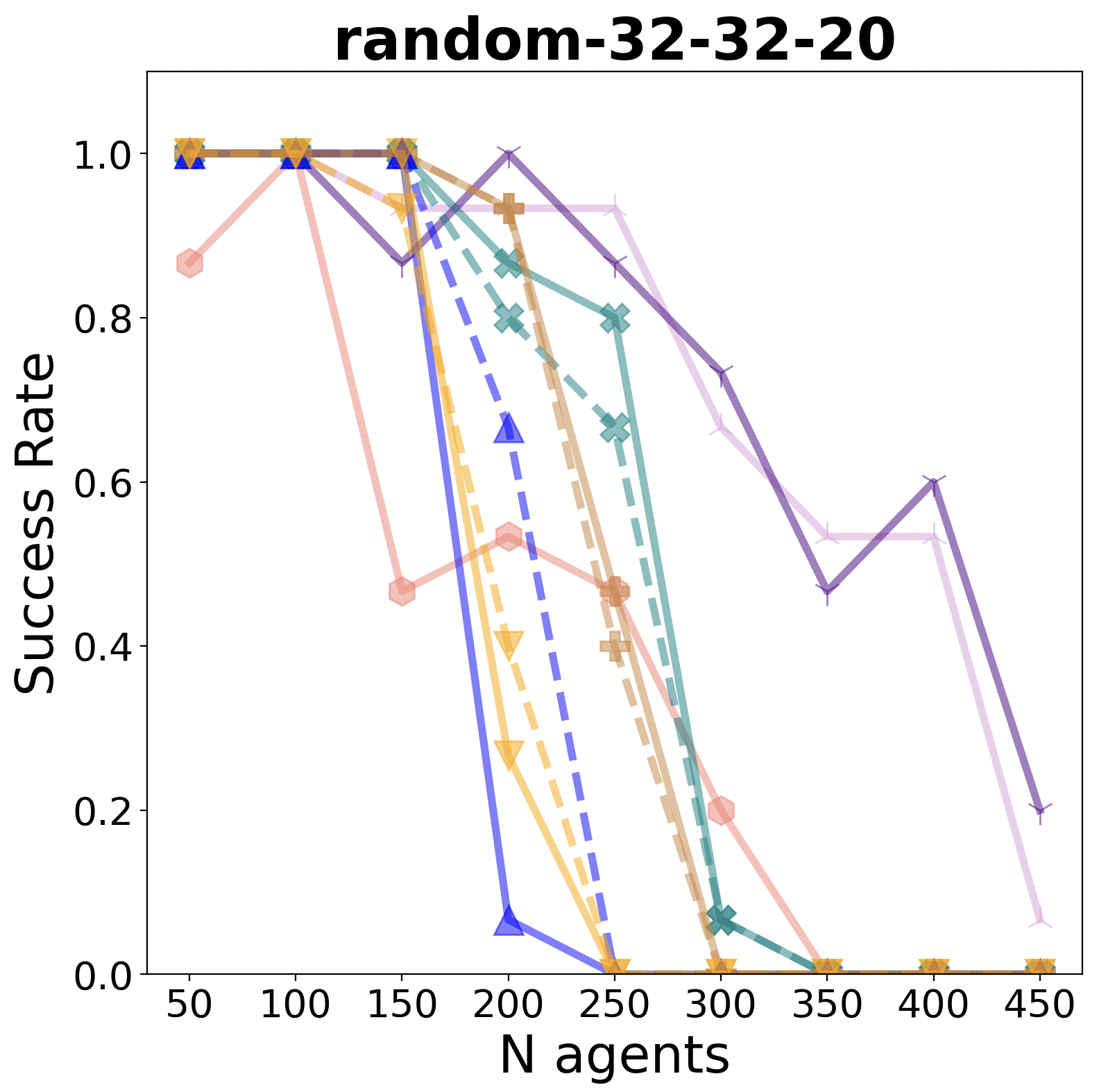}
        \includegraphics[scale=0.15]{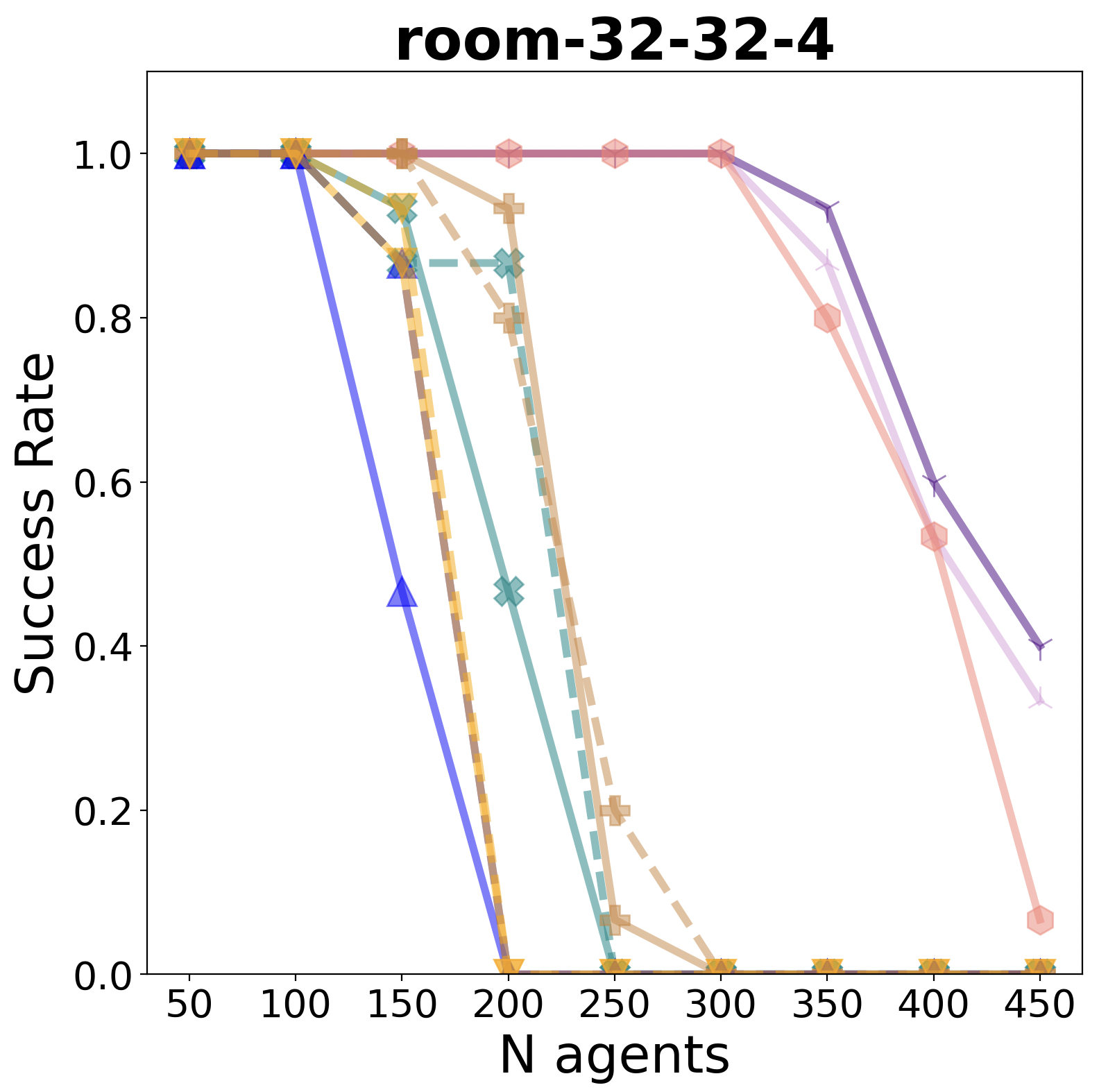}
        \caption{MAPF: Success Rate. Dashed lines - APF-enhanced; Solid lines - no APFs}
        \label{fig:sr}
        % \vspace{-18pt}
    \end{center}
\end{figure*}
Figure \ref{fig:sr} presents the {\em success rate} (SR) of algorithms in different grids, where the SR is the ratio of problems that could be solved within the allocated time limit.  
In many cases, APFs helped to boost the performance, such as for PrP versions. 
But, in some cases, the results were worse with APFs, than without it, such as in LNS2 versions.
Regarding the overall view, LaCAM versions succeded in solving the majority of the problems outperforming others.

\paragraph{Runtime}
\begin{figure*} [!ht]
%\vspace{-8pt}
  % \hspace{-0pt}
 \begin{center}
  \includegraphics[scale=0.15]{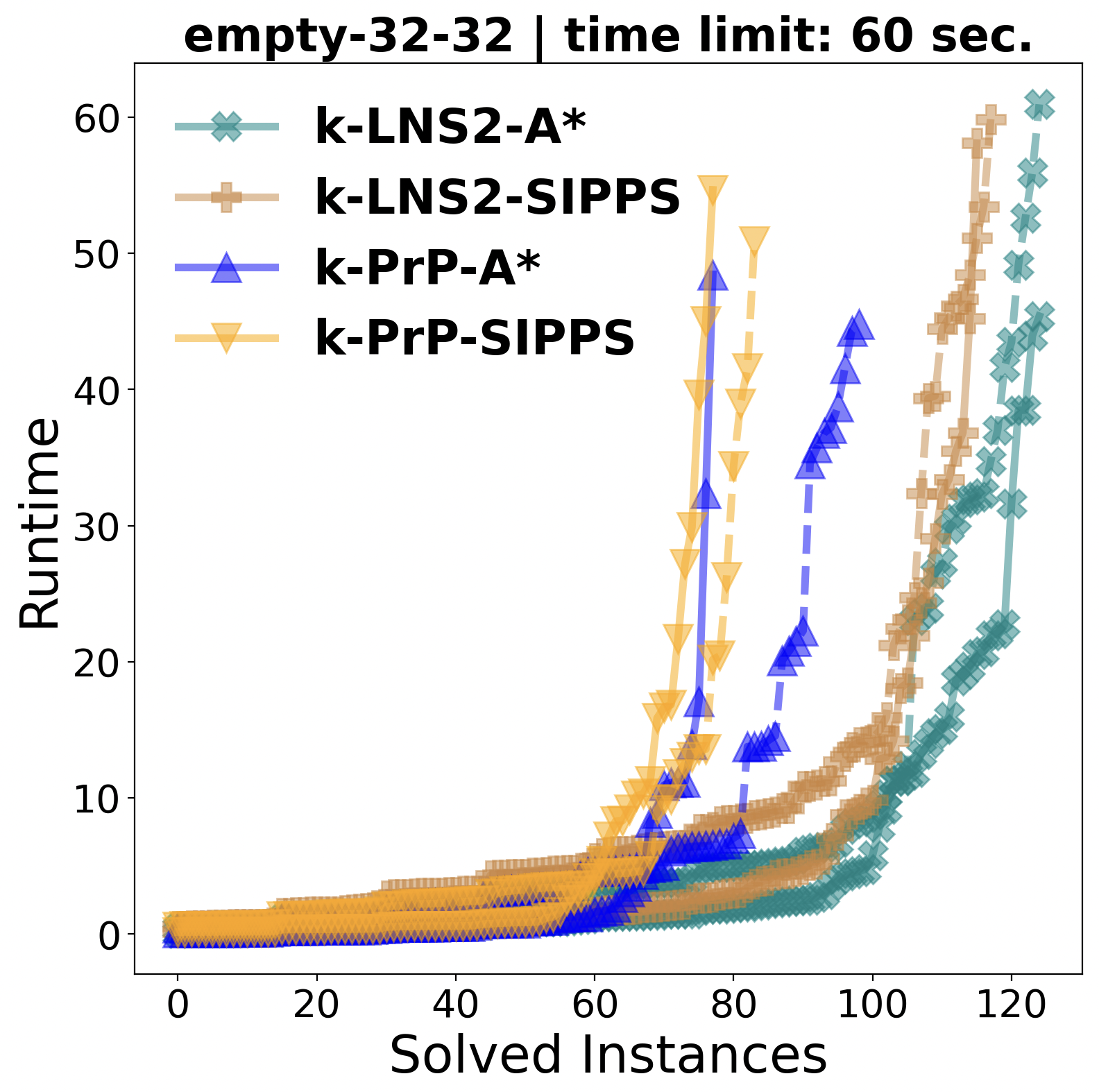}
  \includegraphics[scale=0.15]{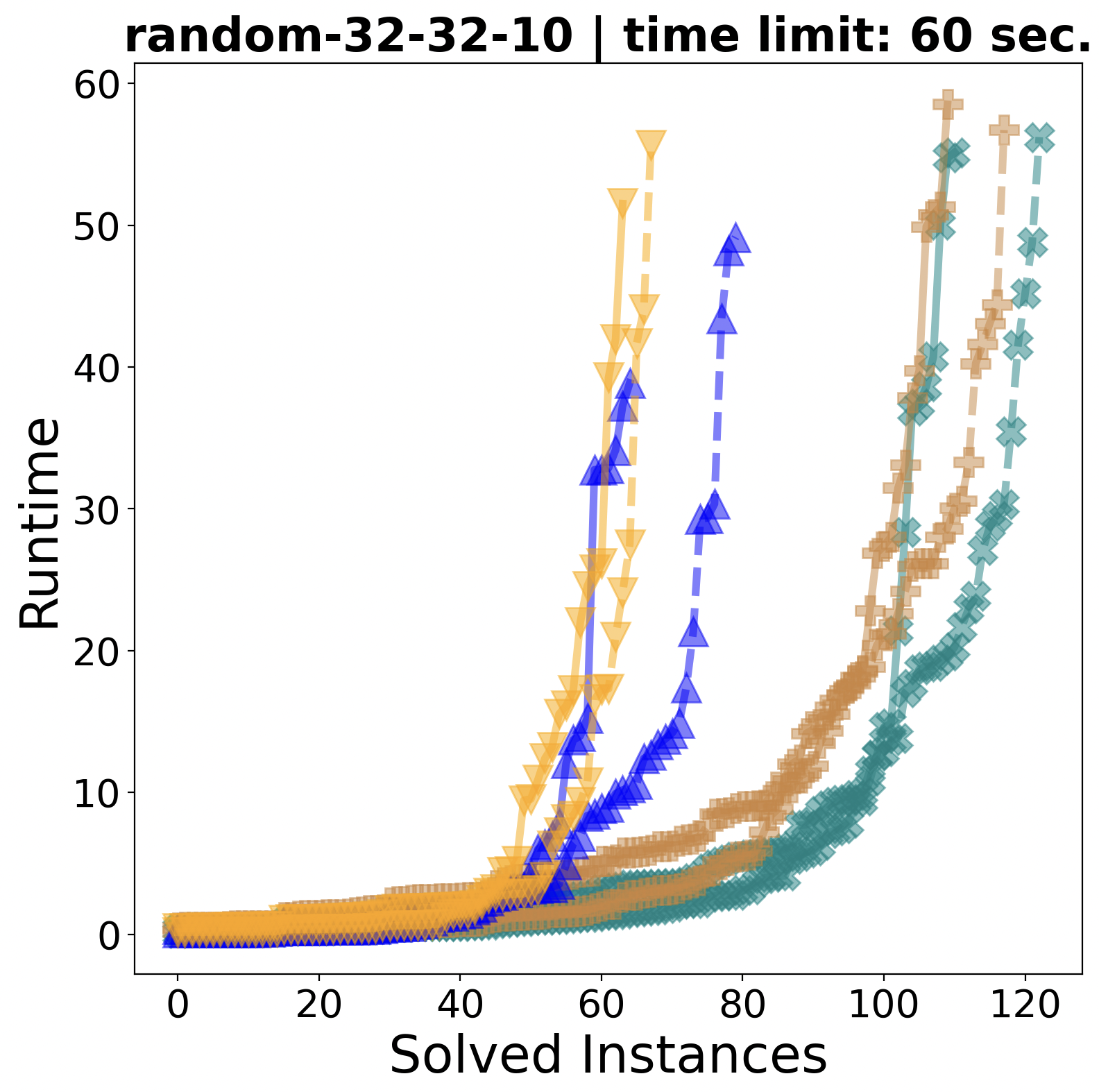}
  \includegraphics[scale=0.15]{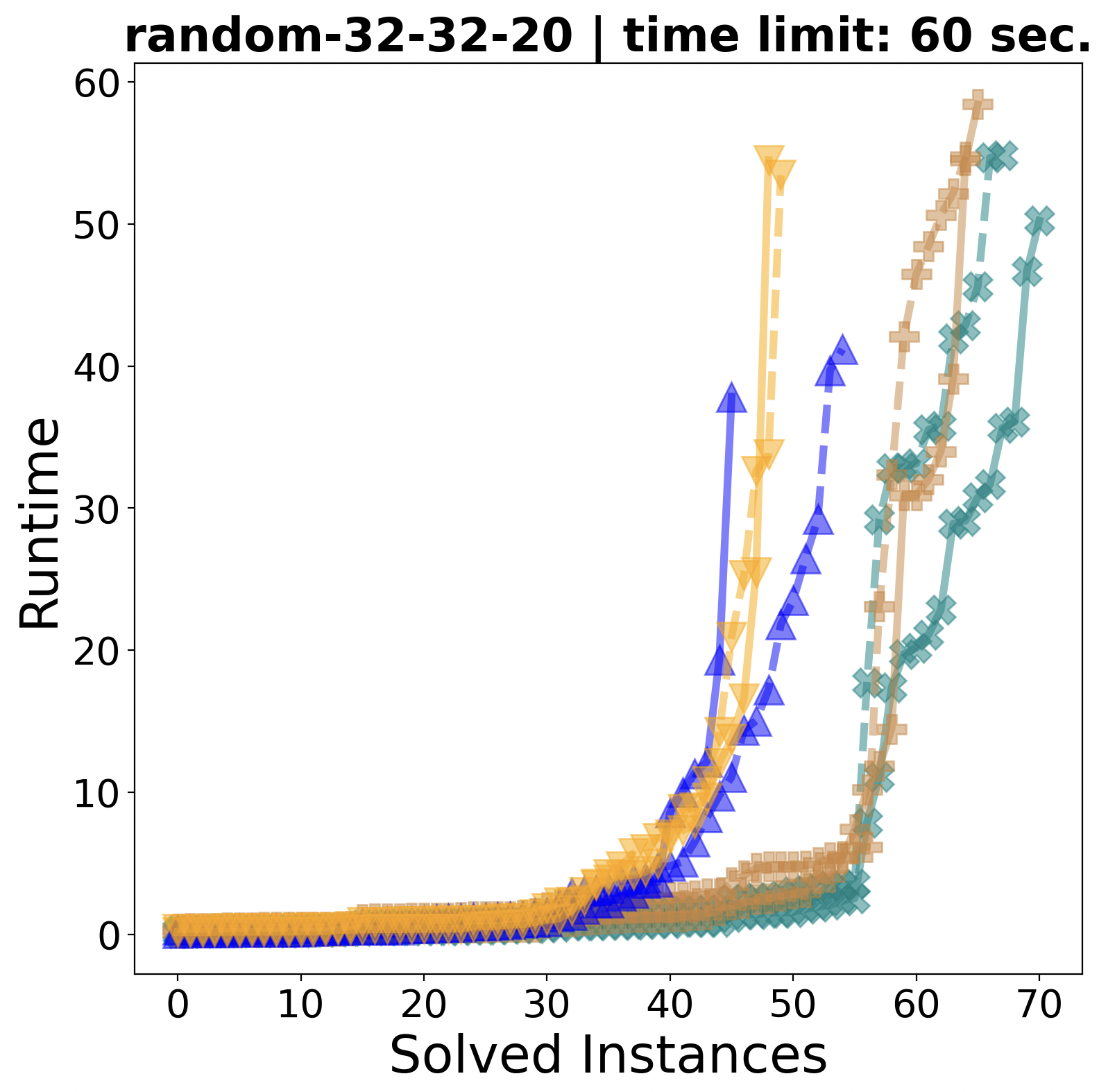}
  \includegraphics[scale=0.15]{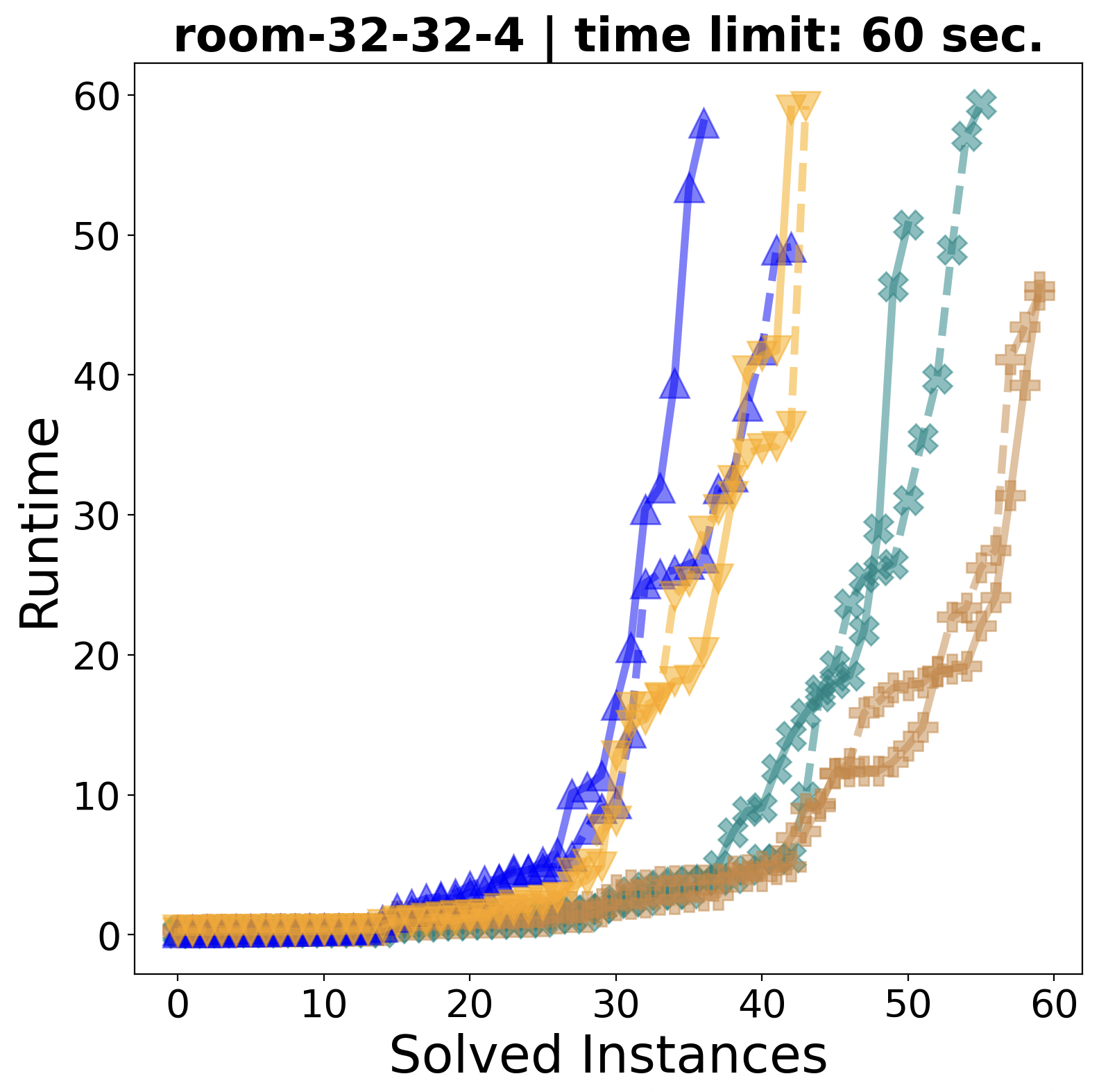}
 \caption{MAPF: Runtime. Plotting the runtime required to solve a given number of instances.}
 \label{fig:rt}
 % \vspace{-18pt}
 \end{center}
 \end{figure*}
Figure \ref{fig:rt} plots the runtime ($y$-axis) required to solve a given number of instances ($x$-axis). This is also known as a ``cactus'' chart. 
The trend is similar to SR metric.
In most cases, the versions of algorithms that used APFs were able to solve more instances in less time than their counterparts. 
For example, in \emph{empty-32-32} the APF-version of PrP with \astar succeeded in solving almost twice as many instances compared to its vanilla variant.
On the other hand, in case of LNS2 in \emph{random-32-32-20}, the results of APF-enhanced versions are the same or even poorer.
% The only exception is in \emph{empty} grids, where LNS2 was faster than PF-LNS2 for the easier instances, i.e., those solved with a smaller runtime. 
% This is because this grid is significantly less constrained than the other grids, and thus in the easier instances resolving conflicts is easier and the APFs mechanism used by APFs is only adding unnecessary overhead. 
% Nevertheless, the advantage of APFs is evident in other more complicated grids. 

\paragraph{RSOC}
\begin{figure*} [!ht]
 \begin{center}
  \includegraphics[scale=0.15]{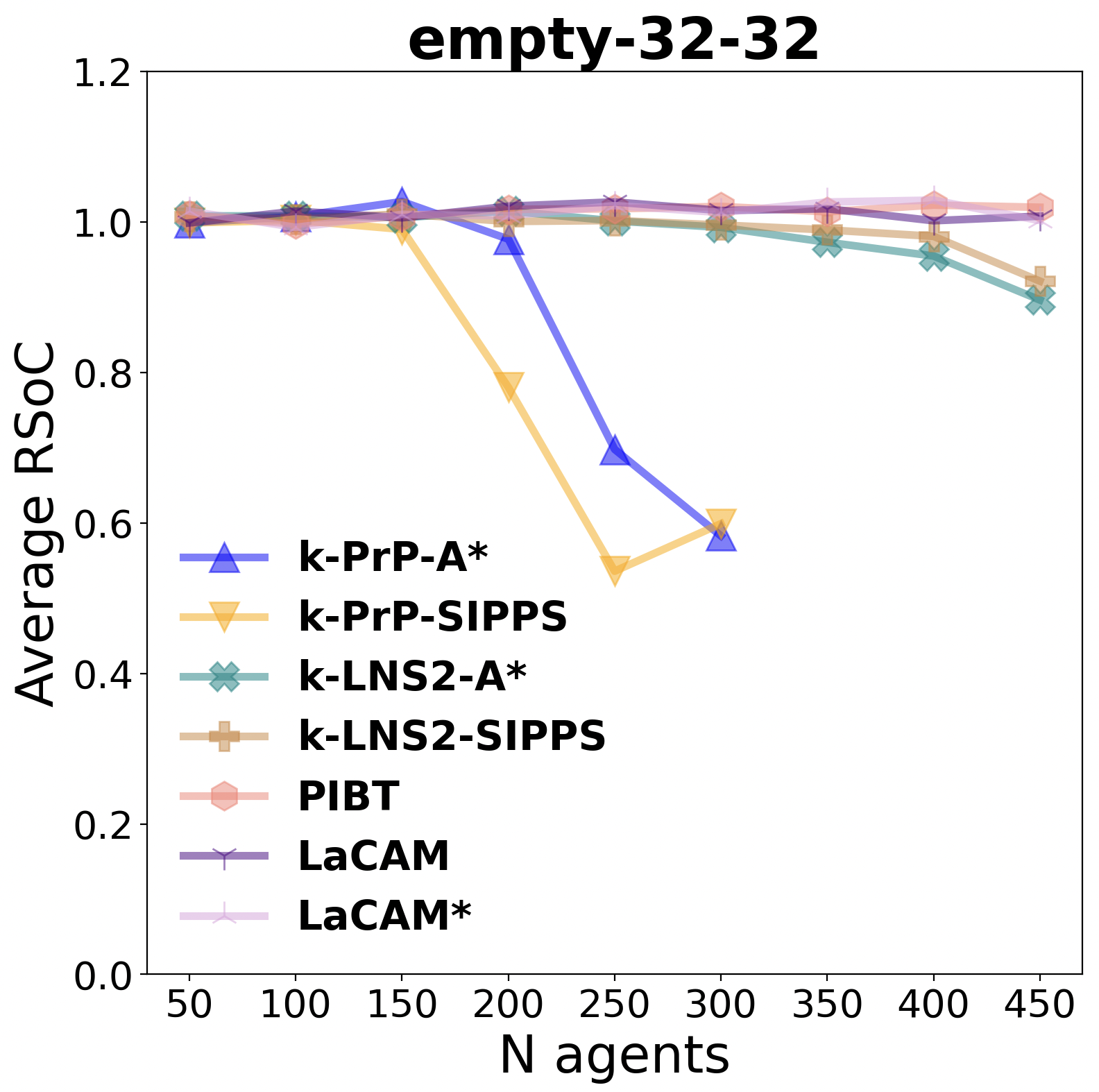}
  \includegraphics[scale=0.15]{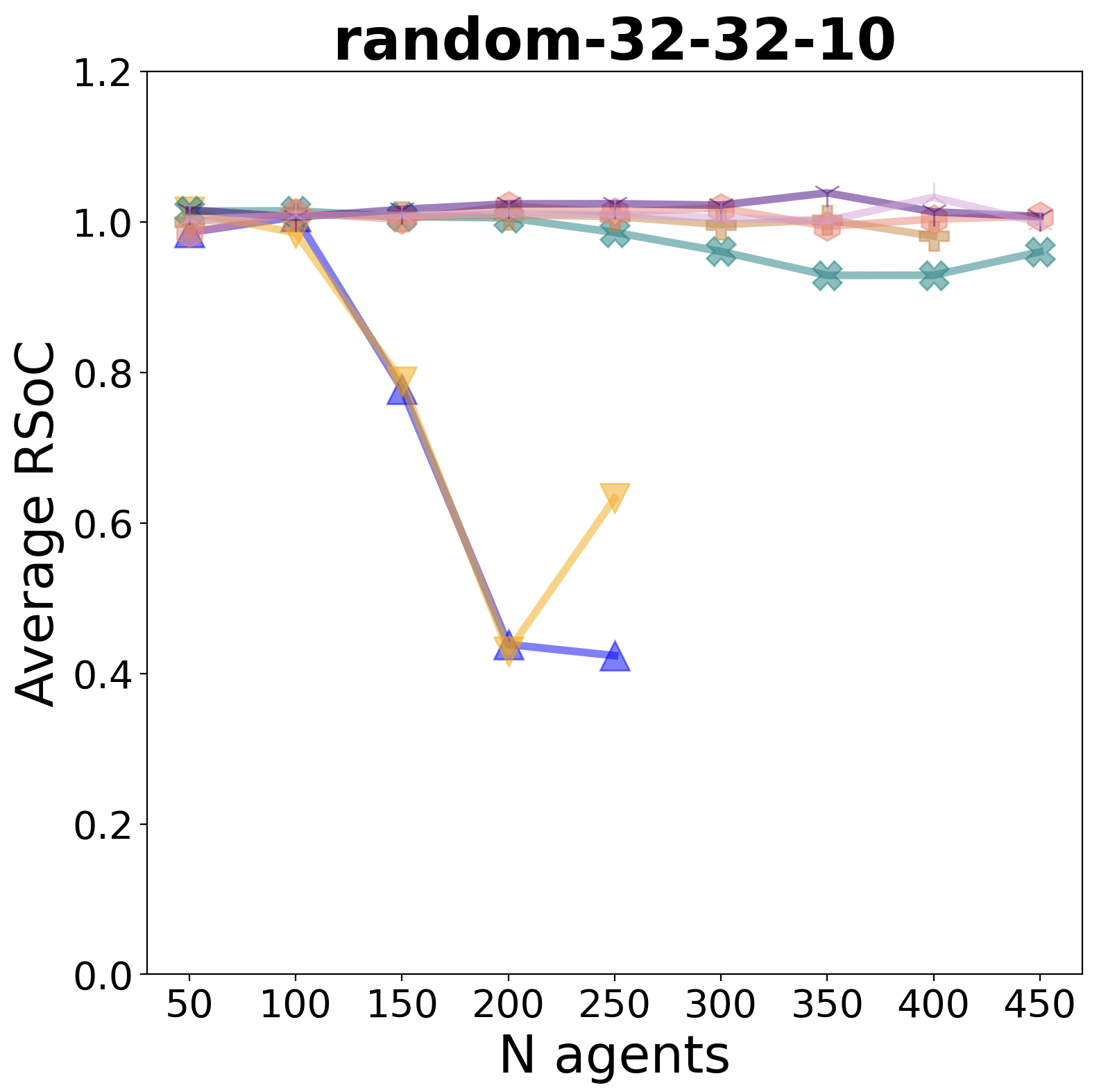}
  \includegraphics[scale=0.15]{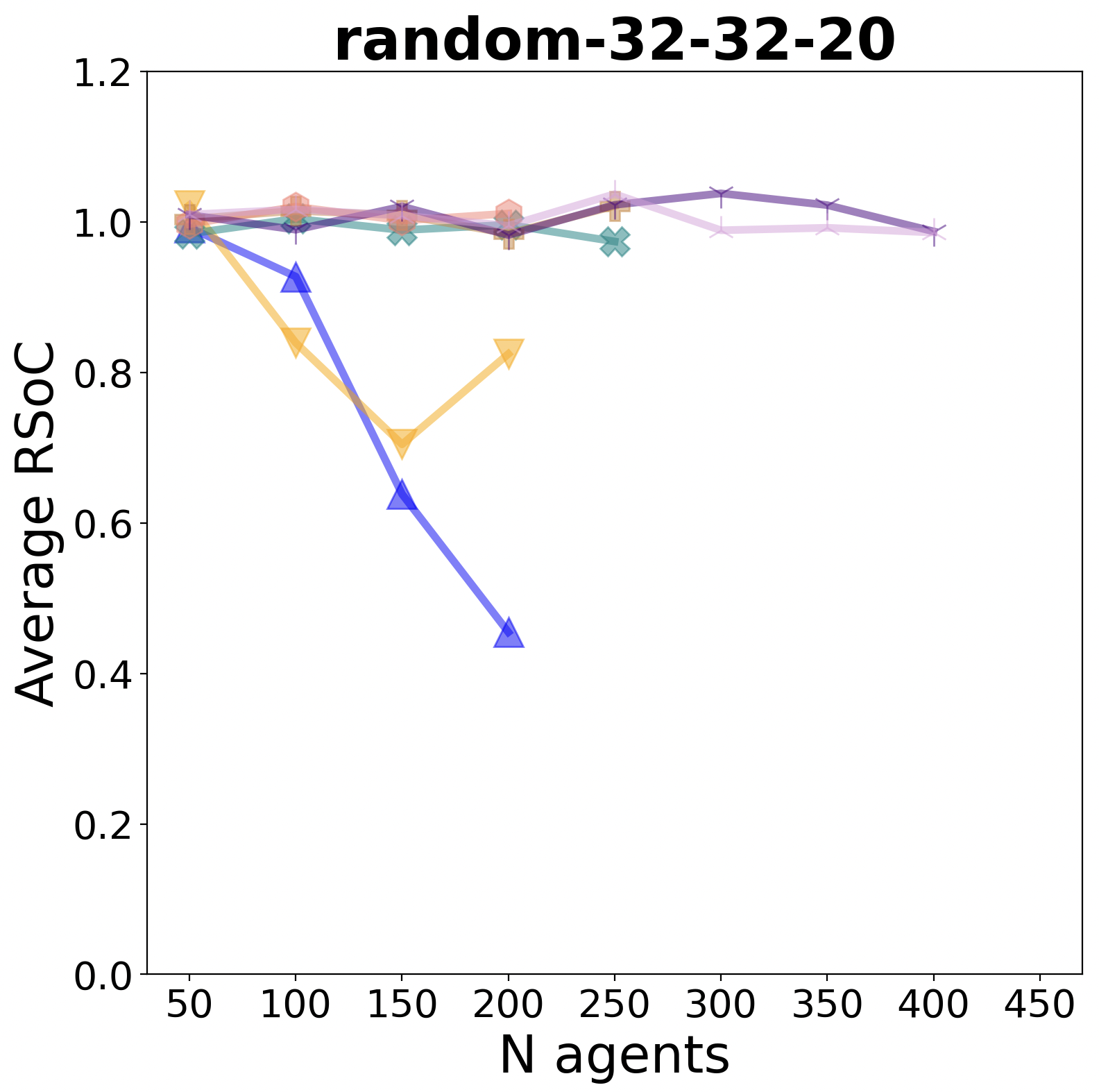}
  \includegraphics[scale=0.15]{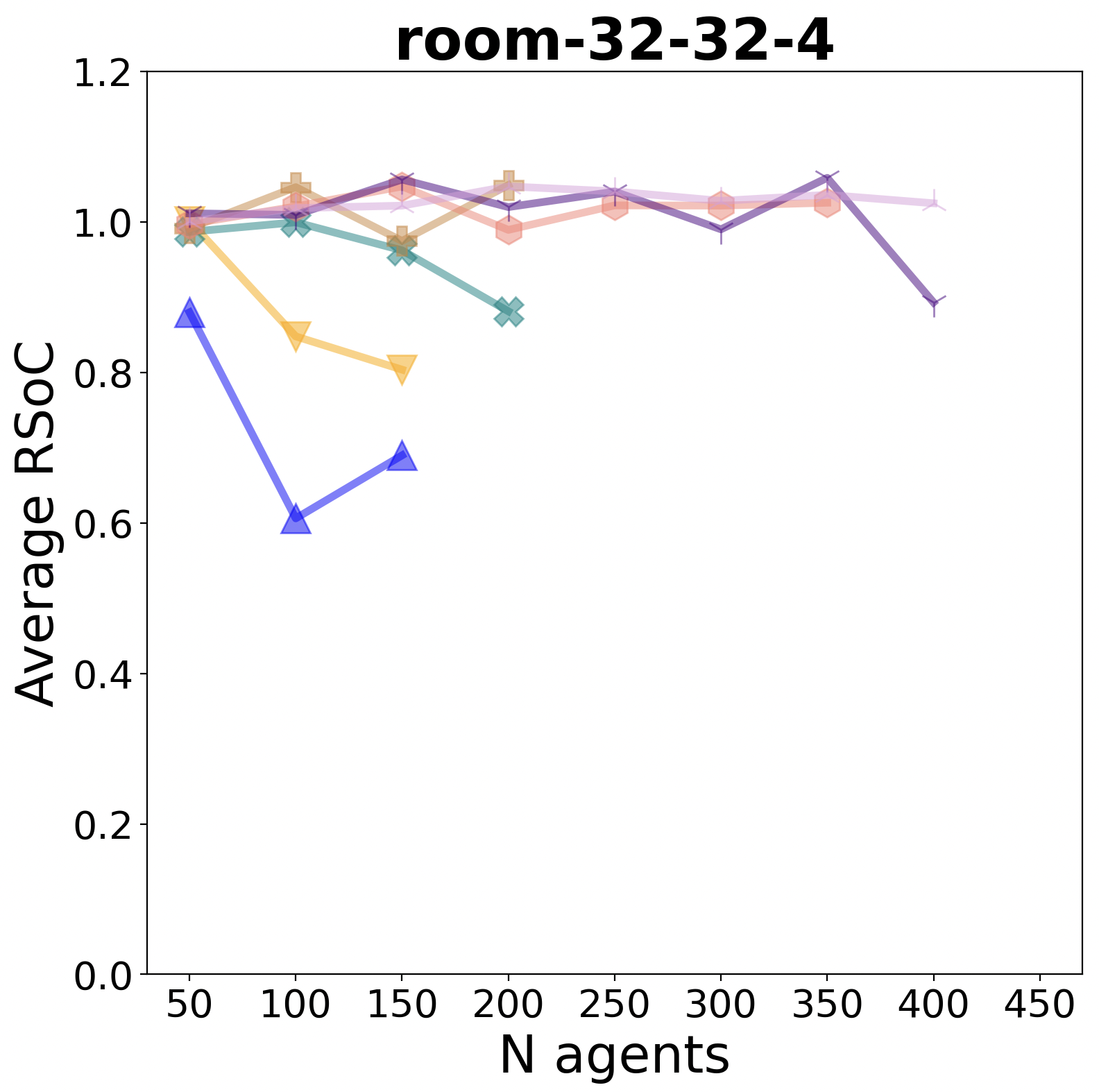}
 \caption{MAPF: RSOC}
 \label{fig:rsoc}
 % \vspace{-18pt}
 \end{center}
 \end{figure*}
Let RSOC denote the ratio between the sum of costs obtained by an APF-enhanced algorithm and the sum of costs obtained by the version of the same algorithm that does not use APFs. 
That is, RSOC for PrP is the sum of the costs of PrP with APFs divided by the sum of the costs of vanilla PrP. 
This is intended to evaluate the potential impact of using APFs on solution cost. 
RSOC smaller than one means APFs reduced the solution cost.
Note that we only compute this ratio for problem instances that were solved by both algorithms. 
Figure \ref{fig:rsoc} presents the RSOC ($y$-axis) of all the benchmark algorithms for each number of agents ($x$-axis) for each of our maps. 
We can see that using APFs does not reduce the solution cost compared to vanilla versions. 
On the contrary, in many cases, it improves the cost. 
Sometimes, the improvement is two-fold, e.g., PrP versions in \emph{random-32-32-10} grids with 200 agents. 
% Empty cells in the table are for cases where one of the algorithms could not solve the relevant problems within the 1-minute time limit.

\end{document}